\pdfoutput=1
\documentclass{article}

\usepackage[utf8]{inputenc} 
\usepackage[T1]{fontenc}    

\usepackage{amsmath} 
\usepackage{amsfonts} 
\usepackage[scale=1.0]{XCharter}

\usepackage[
  paper  = letterpaper,
  top    = 1.0in,
  bottom = 1.0in,
  ]{geometry}

\usepackage[dvipsnames,table]{xcolor}
\definecolor{shadecolor}{gray}{0.9}

\usepackage[final,expansion=alltext]{microtype}
\usepackage[english]{babel}
\usepackage[parfill]{parskip}
\usepackage{afterpage}
\usepackage{framed}

{\endMakeFramed}

\usepackage{lineno}

\usepackage{ragged2e}

\newcounter{parcount}

\usepackage{graphicx}
\usepackage{wrapfig}
\usepackage[labelfont=bf,font=footnotesize,width=.9\textwidth]{caption}
\usepackage[format=hang]{subcaption}

\usepackage{booktabs,multirow,multicol}       

\usepackage[algoruled,algo2e]{algorithm2e}
\usepackage{listings}
\usepackage{fancyvrb}
\fvset{fontsize=\normalsize}

\usepackage[colorlinks,linktoc=all]{hyperref}
\usepackage[all]{hypcap}
\hypersetup{citecolor=MidnightBlue}
\hypersetup{linkcolor=MidnightBlue}
\hypersetup{urlcolor=MidnightBlue}

\usepackage[nameinlink, capitalize]{cleveref}

\usepackage[acronym,smallcaps,nowarn]{glossaries}[=v4.46]

\lstdefinestyle{mystyle}{
    commentstyle=\color{OliveGreen},
    keywordstyle=\color{BurntOrange},
    numberstyle=\tiny\color{black!60},
    stringstyle=\color{MidnightBlue},
    basicstyle=\ttfamily,
    breakatwhitespace=false,
    breaklines=true,
    captionpos=b,
    keepspaces=true,
    numbers=left,
    numbersep=5pt,
    showspaces=false,
    showstringspaces=false,
    showtabs=false,
    tabsize=2
}
\usepackage{amsthm}

\usepackage{centernot}
\usepackage{nicefrac}       
\usepackage{mathtools}
\usepackage{amsbsy}
\usepackage{amstext}
\usepackage{thmtools}
\usepackage{thm-restate}

\begingroup
    \makeatletter
    \@for\theoremstyle:=definition,remark,plain\do{%
        \expandafter\g@addto@macro\csname th@\theoremstyle\endcsname{%
            \addtolength\thm@preskip\parskip
            }%
        }
\endgroup

\crefname{lemma}{lemma}{lemmas}
\Crefname{lemma}{Lemma}{Lemmas}
\crefname{thm}{theorem}{theorems}
\Crefname{thm}{Theorem}{Theorems}
\crefname{prop}{proposition}{propositions}
\Crefname{prop}{Proposition}{Propositions}
\crefname{assumption}{assumption}{assumptions}
\crefname{assumption}{Assumption}{Assumptions}

\usepackage{booktabs}
\makeatletter
\def\adl@drawiv#1#2#3{%
        \hskip.5\tabcolsep
        \xleaders#3{#2.5\@tempdimb #1{1}#2.5\@tempdimb}%
                #2\z@ plus1fil minus1fil\relax
        \hskip.5\tabcolsep}
\newcommand{\cdashlinelr}[1]{%
  \noalign{\vskip\aboverulesep
           \global\let\@dashdrawstore\adl@draw
           \global\let\adl@draw\adl@drawiv}
  \cdashline{#1}
  \noalign{\global\let\adl@draw\@dashdrawstore
           \vskip\belowrulesep}}
\makeatother

\renewcommand{\epsilon}{\varepsilon}

\declaretheorem[style=plain,name=Theorem]{theorem}

\declaretheorem[style=definition,sibling=theorem,name=Example]{example}

\newenvironment{example*}
 {\pushQED{\qed}\example}
 {\popQED\endexample}
\numberwithin{equation}{section}

\DeclarePairedDelimiterX\Set[1]{\lbrace}{\rbrace}%
{  #1 }

\usepackage[utf8]{inputenc} 
\usepackage[T1]{fontenc}    
\usepackage{hyperref}       
\usepackage{url}            
\usepackage{booktabs}       
\usepackage{amsfonts}       
\usepackage{nicefrac}       
\usepackage{microtype}      
\usepackage[table]{xcolor}  
\usepackage{amsmath}
\usepackage{amsthm}
\usepackage{wrapfig,lipsum,booktabs}
\usepackage{multirow}
\usepackage{subcaption}
\usepackage{algorithm}
\usepackage{algpseudocode}
\usepackage{hhline}
\usepackage{bbding}

\definecolor{mydarkblue}{rgb}{0,0.08,0.45}

\hypersetup{ %
    pdftitle={},
    pdfauthor={},
    pdfsubject={},
    pdfkeywords={},
    pdfborder=0 0 0,
    pdfpagemode=UseNone,
    colorlinks=true,
    linkcolor=mydarkblue,
    citecolor=mydarkblue,
    filecolor=mydarkblue,
    urlcolor=mydarkblue,
    pdfview=FitH
}

\usepackage{tcolorbox}
\tcbuselibrary{skins}
\usepackage{afterpage}
\definecolor{linen}{RGB}{250,240,230}
\usepackage{fontawesome5}
\definecolor{linen}{RGB}{250,240,230}
\usepackage{enumitem}
\usepackage{fontawesome5}
\usepackage{caption}
\usepackage{wrapfig,lipsum,booktabs}

\definecolor{FDColor}{RGB}{0,114,178}
\definecolor{FADColor}{RGB}{213,94,0}
\definecolor{OursRowColor}{RGB}{225,240,255}

\crefname{appsec}{appendix}{appendices}
\Crefname{appsec}{Appendix}{Appendices}

\usepackage{appendix}
\usepackage{array}
\newcolumntype{x}[1]{>{\centering\arraybackslash}p{#1pt}}
\newcolumntype{y}[1]{>{\raggedright\arraybackslash}p{#1pt}}

\usepackage{csquotes}
\usepackage[%
minnames=1,maxnames=99,maxcitenames=2,
style=alphabetic,
doi=false,
url=false,
giveninits=true,
hyperref,
natbib,
backend=bibtex,
sorting=nyt,
backref=true
]{biblatex}%
\renewbibmacro{in:}{%
  \ifentrytype{article}{}{\printtext{\bibstring{in}\intitlepunct}}}

\renewbibmacro*{journal}{%
  \iffieldundef{journaltitle}
    {}
    {\printtext[journaltitle]{%
       \printfield[noformat]{journaltitle}%
       \setunit{\subtitlepunct}%
       \printfield[noformat]{journalsubtitle}}}}

\DeclareFieldFormat{sentencecase}{\MakeSentenceCase*{#1}}

\renewbibmacro*{title}{%
  \ifthenelse{\iffieldundef{title}\AND\iffieldundef{subtitle}}
    {}
    {\ifthenelse{\ifentrytype{article}\OR\ifentrytype{inbook}%
      \OR\ifentrytype{incollection}\OR\ifentrytype{inproceedings}%
      \OR\ifentrytype{inreference}\OR\ifentrytype{misc}}
      {\printtext[title]{%
        \printfield[sentencecase]{title}%
        \setunit{\subtitlepunct}%
        \printfield[sentencecase]{subtitle}}}%
      {\printtext[title]{%
        \printfield[titlecase]{title}%
        \setunit{\subtitlepunct}%
        \printfield[titlecase]{subtitle}}}%
     \newunit}%
  \printfield{titleaddon}}

\AtEveryBibitem{%
\ifentrytype{article}{
    \clearfield{urldate}%
    \clearfield{eprint}
    \clearfield{eid}
}{}
\ifentrytype{book}{
    \clearfield{url}%
    \clearfield{urldate}%
    \clearfield{eprint}
}{}
\ifentrytype{collection}{
    \clearfield{url}%
    \clearfield{urldate}%
    \clearfield{eprint}
}{}
\ifentrytype{incollection}{
    \clearfield{url}%
    \clearfield{urldate}%
    \clearfield{eprint}
}{}
}

\AtEveryBibitem{
    \clearfield{pages}
    \clearfield{review}%
    \clearfield{series}
    \clearfield{volume}
    \clearfield{month}
    \clearfield{isbn}
    \clearfield{issn}
    \clearlist{location}
    \clearfield{series}
    \clearlist{publisher}
    \clearname{editor}
}{}

\usepackage{thm-restate}

\crefformat{equation}{(#2#1#3)}
\crefformat{figure}{Figure~#2#1#3}
\crefname{example}{Example}{Examples}
\crefname{lemma}{Lemma}{Lemmas}
\crefname{cor}{Corollary}{Corollaries}
\crefname{theorem}{Theorem}{Theorems}
\crefname{assumption}{Assumption}{Assumptions}

\usepackage{enumitem} 
\usepackage[separate-uncertainty=true,multi-part-units=single]{siunitx} 

\declaretheoremstyle[
spacebelow=\parsep,
    spaceabove=\parsep,
  mdframed={
    backgroundcolor=gray!10!white,     
    hidealllines=true, 
    innertopmargin=8pt, 
    innerbottommargin=4pt, 
    skipabove=8pt,
    skipbelow=10pt,
    nobreak=true
}
]{grayboxed}

\crefname{gassumption}{Assumption}{Assumptions}

\usepackage{xcolor}

\definecolor{WowColor}{rgb}{.75,0,.75}
\definecolor{SubtleColor}{rgb}{0,0,.50}

\newcounter{margincounter}

\usepackage[affil-it]{authblk}

\usepackage{makecell}

\usepackage{multirow}
\usepackage{tikz}
\usetikzlibrary{positioning}
\usetikzlibrary{arrows, automata}
\usetikzlibrary{patterns}

\def\showauthornotes{1}
\ifnum\showauthornotes=1
\newcommand{\Authornote}[2]{{\sf\small\color{blue}{[#1: #2]}}}
\else
\newcommand{\Authornote}[2]{}
\fi

\title{Efficient Fréchet Distributional Decoder Alignment for Latent Generative Models}

\author{
  {\bfseries\upshape
  Xianghong Fang$^{1,*}$
  \quad
  Jiajun Zhu$^{2,*}$
  \quad
  Esma Aimeur$^{2}$
  } \\ \vspace{-5pt}
  {\bfseries\upshape
  Dehan Kong$^{1}$
  \quad
  Tim G. J. Rudner$^{1,3}$
  } \\
  {\upshape
  \vspace*{10pt}
  $^1$University of Toronto \quad
  $^2$Université de Montréal \quad
  $^3$Vijil 
  } \\
  {\upshape
  \vspace*{15pt}
    \begin{center}
    \href{https://sunset-clouds.github.io/Frechet-Distributional-Decoder-Alignment/}{\faGlobe\enspace Website}
    \quad
    \href{https://github.com/sunset-clouds/Frechet-Distributional-Decoder-Alignment}{\faGithub\enspace Code \& Models}
    \end{center}
  }
}
\date{}

\begin{document}

\maketitle

\begingroup
\renewcommand\thefootnote{}
\footnotetext{
$^{*}$ Equal contribution.
}
\endgroup

\begin{abstract}
Fréchet Distributional (FD) objectives provide an effective distribution-level signal for post-training generative models. For latent generative models, however, existing FD-based post-training methods optimize only the latent generator while keeping the decoder frozen: the decoder is trained on reconstruction latents obtained from real images but deployed on generator-produced latents at generation time. When these latent distributions differ, this \textbf{decoder train-generation mismatch} means that the decoder is deployed on inputs drawn from a distribution it was not directly trained on. We introduce \textbf{Fréchet Distributional Decoder Alignment (FDDA)}, an efficient post-training method that freezes the latent generator and directly optimizes the decoder on generation-time latents using an FD objective. A single epoch of FDDA consistently improves gFID and gFDr$^6$ across diverse flow-based and autoregressive latent generative models. Across all 11 evaluated models, FDDA reduces gFDr$^6$ by 34.0-59.6\% from the original pretrained models and provides a further 10.2-57.2\% reduction after generator-side FD post-training. FDDA also reduces total wall-clock time by factors of $3.6\times$-$328.7\times$ relative to generator-side FD post-training across these models. These results show that generator-side post-training alone does not exhaust the benefits of FD optimization, and that further gains can be obtained by adapting the decoder at substantially lower cost.
\end{abstract}


\section{Introduction}
\label{sec:intro}

Post-training has become an effective way to further improve pretrained generative models without retraining them from scratch~\citep{Xu2023ImageRewardLA,Black2024Training,Luo2024David,Fang2026GenerationAwareRecon,Lin2026Continuous}. In particular, Fréchet Distributional (FD) objectives match the representation-space statistics of generated and real samples, providing a distribution-level objective for post-training~\citep{Yang2026RepresentationFL}. Recent FD-based post-training methods have demonstrated substantial improvements in generation quality across diverse generative model families~\citep{Yang2026RepresentationFL,Zhang2026Frechetdistributional,Gao2026Advfd}. Yet in latent generative models, these methods optimize only the generator; the decoder has not been studied as a target for distribution-level post-training~\citep{Yang2026RepresentationFL,Zhang2026Frechetdistributional}.

This raises a natural question:
\vspace*{1pt}
\begin{tcolorbox}[
  enhanced,
  colback=blue!4,
  colframe=blue!55!black,
  leftrule=3pt,
  toprule=0.4pt,
  bottomrule=0.4pt,
  rightrule=0.4pt,
  arc=2pt,
  boxsep=1.5pt,
  left=8pt, right=8pt, top=2pt, bottom=2pt,
  width=\linewidth,
  before skip=2pt,
  after skip=5pt,
]
\begin{center}
    {\it Does improving the generator exhaust the potential of distribution-level post-training?}
\end{center}
\end{tcolorbox}
\vspace*{-1pt}

For latent generative models, two observations suggest otherwise.
First, generation relies on two components:
Alatent generator that produces a generation-time latent $z_g \sim p_g(z)$, and a decoder that maps it to an image~\citep{Oord2017NeuralDR,Esser2021TamingTF,Rombach2022HighResolutionIS,Sun2024Autoregressive}. 
Since both components affect final image quality, the decoder can also be optimized with a distribution-level objective. Second, the decoder faces a \textbf{train-generation mismatch}:
It is typically trained on reconstruction latents $z_r \sim p_r(z)$ obtained by encoding real images~\citep{Oord2017NeuralDR,Esser2021TamingTF,Rombach2022HighResolutionIS}, yet deployed on generation-time latents $z_g$ produced by the generator.
When $p_r(z)$ and $p_g(z)$ differ, the decoder receives generation-time inputs that it was not directly trained on~\citep{Qiu2025ImageTN,Fang2026GenerationAwareRecon}.
Generator-side post-training does not directly adapt the decoder to its generation-time inputs: it updates the generator and hence $p_g(z)$, while the decoder remains optimized primarily on $p_r(z)$.

\begin{figure}[t]
\vspace{-3mm}
\centering
\captionsetup[subfloat]{skip=3pt}
\setlength{\tabcolsep}{0pt}
\newcommand{\panelsep}{%
  \par
  \vskip 0.2ex
  \hrule height 0.3pt
  \vskip 0.4ex
}

\subfloat[Generator-side vs.\ decoder-side FD post-training]{
\label{fig:FD comparison}
\begin{tabular}{c@{\hspace{0.02\linewidth}}c}
\includegraphics[width=0.48\linewidth]{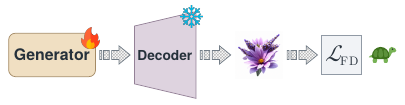} &
\includegraphics[width=0.48\linewidth]{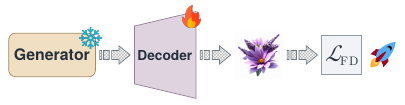} \\[1pt]
{\footnotesize Generator-side FD post-training} &
{\footnotesize Decoder-side FD post-training}
\end{tabular}}
\panelsep

\subfloat[FDDA gains before (top) and after (bottom) generator-side FD post-training]{
\label{fig:performance gains}
\begin{tabular}{c@{\hspace{0.01\linewidth}}c@{\hspace{0.01\linewidth}}c@{\hspace{0.01\linewidth}}c}
\includegraphics[width=0.24\linewidth]{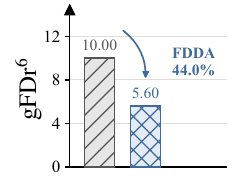} &
\includegraphics[width=0.24\linewidth]{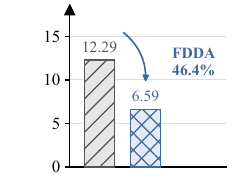} &
\includegraphics[width=0.24\linewidth]{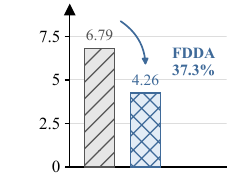} &
\includegraphics[width=0.24\linewidth]{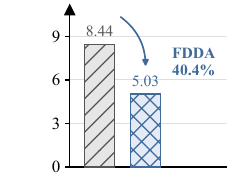} \\[-2pt]
\includegraphics[width=0.24\linewidth]{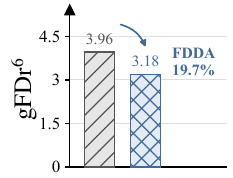} &
\includegraphics[width=0.24\linewidth]{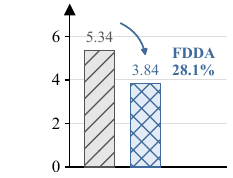} &
\includegraphics[width=0.24\linewidth]{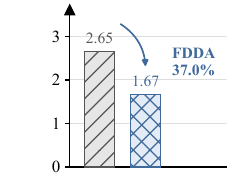} &
\includegraphics[width=0.24\linewidth]{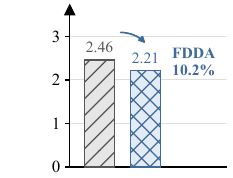} \\[1pt]
{\footnotesize LlamaGen-L} & {\footnotesize GigaTok-S-S} &
{\footnotesize VAR-$d24$} & {\footnotesize iMF-XL/2}
\end{tabular}}
\panelsep

\subfloat[Wall-clock cost of generator-side vs.\ decoder-side FD post-training]{
\label{fig:training cost}
\begin{tabular}{c@{\hspace{0.01\linewidth}}c@{\hspace{0.01\linewidth}}c@{\hspace{0.01\linewidth}}c}
\includegraphics[width=0.24\linewidth]{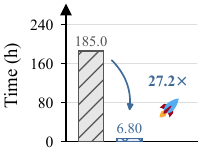} &
\includegraphics[width=0.24\linewidth]{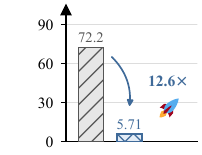} &
\includegraphics[width=0.24\linewidth]{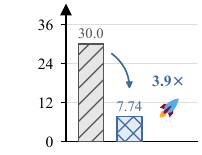} &
\includegraphics[width=0.24\linewidth]{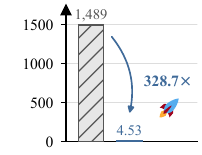} \\[1pt]
{\footnotesize LlamaGen-L} & {\footnotesize GigaTok-S-S} &
{\footnotesize VAR-$d24$} & {\footnotesize iMF-XL/2}
\end{tabular}}

\vspace{-1ex}
\caption{\small{
\textbf{Decoder-side FD post-training with FDDA.}
\textbf{(a)} Existing methods optimize the latent generator with a frozen decoder, whereas FDDA freezes the generator and optimizes the decoder on generation-time latents.
\textbf{(b)} FDDA provides complementary gains. Top row: FDDA applied to released checkpoints before generator-side FD post-training. Bottom row: FDDA applied after generator-side FD post-training. Gray and blue bars denote gFDr$^6$ before and after FDDA, and annotations report the relative reduction.
\textbf{(c)} Decoder-side FD post-training substantially reduces training cost. Gray (left) and blue (right) bars denote total wall-clock hours on a single H100 GPU for generator-side and decoder-side FD post-training, and annotations show the reduction factor.}}
\label{fig:teaser}
\vspace{-2ex}
\end{figure}

To address this mismatch, we introduce \textbf{Fréchet Distributional Decoder Alignment (FDDA)}, which extends FD-based post-training to the decoder, as illustrated in Figure~\ref{fig:FD comparison}. FDDA freezes the pretrained latent generator and optimizes the decoder on generation-time latents $z_g \sim p_g(z)$ using an FD objective on the generated images. In this way, FDDA adapts the decoder to the latent distribution it receives during generation while leaving the pretrained generator unchanged, providing a decoder-side complement to generator-side FD post-training.

FDDA improves generation quality across both flow-based and autoregressive latent generative models. Across all 11 evaluated configurations,  FDDA improves both gFID~\citep{Heusel2017GANsTB} and gFDr$^6$~\citep{Yang2026RepresentationFL} when applied to the original pretrained models. For the four configurations shown in Figure~\ref{fig:performance gains}, gFDr$^6$ decreases by 44.0\% on LlamaGen-L~\citep{Sun2024Autoregressive}, 46.4\% on GigaTok-S-S~\citep{Xiong2025GigaTokSV}, 37.3\% on VAR-$d24$~\citep{Tian2024VAR}, and 40.4\% on iMF-XL/2~\citep{Geng2025ImprovedMF}. The gains remain after generator-side FD post-training: FDDA further reduces gFDr$^6$ by 19.7\%, 28.1\%, 37.0\%, and 10.2\% on these four configurations, respectively, while improving gFDr$^6$ across all 11 evaluated configurations. FDDA also reduces the wall-clock cost of FD post-training. As shown in Figure~\ref{fig:training cost}, its total wall-clock time is lower by factors of $27.2\times$, $12.6\times$, $3.9\times$, and $328.7\times$ for the LlamaGen, GigaTok, VAR, iMF configurations, respectively. These results show that generator-side post-training alone does not exhaust the benefits of FD optimization, \mbox{as further gains can still be obtained by adapting the decoder at substantially lower cost.}

Our contributions are threefold:
\begin{enumerate}[topsep=0pt, align=left, leftmargin=15pt, labelindent=0pt,
listparindent=\parindent, labelwidth=0pt, itemindent=!, itemsep=3pt, parsep=0pt]
\item We introduce \textbf{Fréchet Distributional Decoder Alignment (FDDA)}, which freezes the latent generator and optimizes the decoder on generation-time latents with an FD objective, directly addressing the decoder train-generation mismatch.
\item We show that a single epoch of FDDA improves gFID and gFDr$^6$ across all 11 evaluated flow-based and autoregressive latent generative models, while requiring substantially less total wall-clock time than generator-side FD post-training.
\item We show that FDDA provides further gFDr$^6$ improvements after generator-side FD post-training across all evaluated models, indicating that generator-side optimization alone does not exhaust the benefits of FD post-training.
\end{enumerate}
\section{Related Work}
\label{sec:related_work}
\vspace{-1ex}
\paragraph{Post-training generative models.} Post-training has emerged as an effective approach for improving pretrained generative models with objectives that differ from those used during pretraining. Prior work has optimized diffusion models toward downstream rewards or human preferences. Reward-based approaches include direct reward feedback~\citep{Xu2023ImageRewardLA}, reinforcement-learning-based fine-tuning~\citep{Black2024Training,Fan2023DPOKRL}, direct back-propagation through differentiable rewards~\citep{Clark2024Directly}, and scalable reward optimization~\citep{Deng2024PRDPPR}. Preference-based methods instead directly fine-tune pretrained diffusion models from pairwise feedback~\citep{Wallace2024DiffusionDPO,Yang2024UsingHF}. Other work has explored adversarial objectives~\citep{Lin2026Continuous} and distribution-level objectives~\citep{Yang2026RepresentationFL,Zhang2026Frechetdistributional,Gao2026Advfd} for post-training pretrained generators. Our work follows this general direction, but applies distribution-level post-training to the decoder rather than further optimizing the latent generator.

\vspace{-1ex}
\paragraph{Fréchet distributional post-training.} Fréchet Inception Distance (FID) was introduced as an evaluation metric for generative models~\citep{Heusel2017GANsTB}. Earlier work explored feature-moment matching and direct optimization of Fréchet-style objectives~\citep{Mathiasen2021BackpropagatingTF,Doan2020ImageGV}, but these methods were limited by batch-scale statistics. FD loss~\citep{Yang2026RepresentationFL} makes large-population FD optimization practical by decoupling the statistics used for FD estimation from the batch used for gradient computation, making it suitable for post-training pretrained visual generators. Subsequent work extends FD-based post-training to autoregressive image generators~\citep{Zhang2026Frechetdistributional} and introduces adversarially learned representation spaces for FD optimization~\citep{Gao2026Advfd}. Existing scalable FD-based post-training methods, however, optimize the generator while keeping the decoder fixed in latent generative models. FDDA instead applies FD-based post-training directly to the decoder on generation-time latents.

\vspace{-1ex}
\paragraph{Decoder alignment with generation-time latents.} Latent generative models typically train their tokenizer or autoencoder separately from the downstream generator~\citep{Oord2017NeuralDR,Esser2021TamingTF,Rombach2022HighResolutionIS}. The decoder is therefore trained on reconstruction latents but used on generator-produced latents at generation time, resulting in a decoder train-generation mismatch~\citep{Qiu2025ImageTN,Fang2026GenerationAwareRecon}. Prior work has reduced this mismatch by adapting the decoder using latents that more closely resemble those encountered during generation. RobusTok~\citep{Qiu2025ImageTN} performs generator-aware decoder post-training and, for autoregressive generators, introduces a preservation ratio that retains part of the reconstruction tokens during sampling. Generation-aware reconstruction~\citep{Fang2026GenerationAwareRecon} constructs a reconstruction-to-generation trajectory and adapts the decoder on intermediate generative latents while preserving correspondence with source images. These methods use paired supervision and retain per-sample reconstruction targets during decoder adaptation. FDDA instead optimizes the decoder directly on fully generated latents $z_g \sim p_g(z)$ with an FD objective while keeping the latent generator fixed. Its distribution-level objective does not require per-sample correspondence between generated and real images.

\section{Preliminaries and Problem Setup}
\label{sec:preliminaries}

\subsection{Latent Generative Models}
\label{sec:prelim_latent}

Latent generative models typically follow a two-stage training paradigm~\citep{Oord2017NeuralDR,Esser2021TamingTF,Rombach2022HighResolutionIS,Sun2024Autoregressive}. First, a tokenizer or autoencoder learns a latent representation of images. Given $x \sim p_{\mathrm{data}}(x)$, the encoder produces a reconstruction latent $z_r = \mathcal{E}_{\theta}(x) \sim p_r(z)$, from which the decoder $\mathcal{D}_{\phi}$ is trained to reconstruct $x$. Here, $p_r(z)$ denotes the encoder-induced distribution over decoder inputs. A latent generator is then trained over this latent space and, at generation time, produces latents $z_g \sim p_g(z)$ that are decoded as $x_g = \mathcal{D}_{\phi}(z_g)$.

We use $z_g$ to denote the generation-time input to the decoder regardless of the generator type. For diffusion- and flow-based latent generators, $z_g$ is obtained from the final state of the sampling process~\citep{Rombach2022HighResolutionIS,Peebles2022ScalableDM,Geng2025MeanFF,Geng2025ImprovedMF}. For autoregressive latent generators, the generator first samples discrete tokens, which are then converted through the tokenizer codebook into generation-time latents $z_g$~\citep{Esser2021TamingTF,Sun2024Autoregressive,Xiong2025GigaTokSV,Yu2024AnII,Tian2024VAR}. We use $p_g(z)$ to denote the resulting distribution over decoder inputs in both settings.

The decoder is therefore trained on reconstruction latents $z_r \sim p_r(z)$ but used on generation-time latents $z_g \sim p_g(z)$. When $p_r(z)$ and $p_g(z)$ differ, the decoder is applied to inputs drawn from a distribution on which it was not directly trained. We refer to this as the \textbf{decoder train-generation mismatch}~\citep{Qiu2025ImageTN,Fang2026GenerationAwareRecon}.

\subsection{Fr\'echet Distributional Objectives}
\label{sec:prelim_fd}

Fr\'echet Distance (FD) measures distributional discrepancy by comparing the first- and second-order statistics of real and generated samples in a representation space~\citep{Heusel2017GANsTB,Yang2026RepresentationFL}. Let $\psi(\cdot)$ denote a representation model. For real images $x \sim p_{\mathrm{data}}(x)$ and generated images $x_g$, let $\mu_r$ and $\Sigma_r$ denote the mean and covariance of $\psi(x)$, and let $\mu_g$ and $\Sigma_g$ denote the corresponding statistics of $\psi(x_g)$. The Fr\'echet distance is
\begin{equation}
    \operatorname{FD}_{\psi}
    =
    \|\mu_r - \mu_g\|_2^2
    +
    \operatorname{Tr}
    \left(
        \Sigma_r
        +
        \Sigma_g
        -
        2
        \left(
            \Sigma_r^{1/2}
            \Sigma_g
            \Sigma_r^{1/2}
        \right)^{1/2}
    \right).
    \label{eq:fd}
\end{equation}
The choice of $\psi$ specifies the feature space of the FD objective: Inception-v3~\citep{szegedy2016rethinking} recovers the standard Fr\'echet Inception Distance (FID)~\citep{Heusel2017GANsTB}, while other pretrained representations define FD objectives in their respective feature spaces~\citep{Yang2026RepresentationFL}.

Directly optimizing FD is challenging because estimating its population statistics requires substantially more samples than are available in a typical minibatch. FD-loss~\citep{Yang2026RepresentationFL} addresses this issue by decoupling the population used for FD estimation from the minibatch used for gradient computation, making large-population FD optimization practical for post-training diffusion- and flow-based generative models. Zhang et al.~\citep{Zhang2026Frechetdistributional} extend FD post-training to discrete autoregressive generators via rollout-based optimization and straight-through estimation. AdvFD~\citep{Gao2026Advfd} further learns the feature space adversarially to expose remaining discrepancies between real and generated distributions.

Existing FD-based post-training methods are \emph{generator-side}, optimizing the generator while keeping the decoder fixed. In latent generative models, they therefore do not directly adapt the decoder to its generation-time inputs. We instead apply FD optimization to the decoder on generation-time latents, as described in the next section.

\section{Fr\'echet Distributional Decoder Alignment}
\label{sec:fdda}

\subsection{Decoder-Side FD Post-Training}
\label{sec:fdda_decoder}

We introduce \textbf{Fr\'echet Distributional Decoder Alignment (FDDA)}, which extends FD post-training from the generator to the decoder. Given a pretrained latent generator with generation-time latent distribution $p_g(z)$, FDDA freezes the generator and updates only the decoder $\mathcal{D}_{\phi}$. Specifically, generation-time latents $z_g \sim p_g(z)$ are sampled from the frozen generator and decoded as $x_g = \mathcal{D}_{\phi}(z_g)$. This defines the generated image distribution $p_{\phi}^{x} = (\mathcal{D}_{\phi})_{\#}p_g$, where $p_{\phi}^{x}$ is the distribution of generated images obtained by decoding $z_g \sim p_g(z)$. FDDA optimizes the decoder by matching $p_{\phi}^{x}$ to the real data distribution:
\begin{equation}
    \phi^{\star}
      =
    \arg\min_{\phi}
    \operatorname{FD}_{\psi}
    \left(
        p_{\mathrm{data}},
        p_{\phi}^{x}
    \right),
    \qquad
    p_g \text{ fixed}.
    \label{eq:fdda_objective}
\end{equation}
Generator-side FD post-training updates the generator and hence changes $p_g(z)$. FDDA instead keeps $p_g(z)$ fixed and trains the decoder directly on samples from the latent distribution encountered at generation time, reducing the mismatch between decoder training and generation.

FDDA uses distribution-level supervision and requires neither reconstruction targets nor per-sample correspondence between generated and real images. This differs from paired decoder adaptation methods~\citep{Qiu2025ImageTN,Fang2026GenerationAwareRecon}, which retain sample-level reconstruction supervision. FDDA directly trains the decoder on fully generated latents $z_g \sim p_g(z)$. The formulation is the same for diffusion- and flow-based and autoregressive latent generators; only the procedure used to obtain $z_g$ changes. Gradients in FDDA pass only through the decoder. The latent generator and its sampling process are not part of the backward pass, which avoids the cost of differentiating through multi-step sampling or autoregressive rollouts during generator-side FD post-training.

\subsection{Multi-Representation FD Optimization}
\label{sec:fdda_optimization}

FD can be computed in different representation spaces, each capturing different aspects of distributional discrepancy. Following FD-loss~\citep{Yang2026RepresentationFL}, we adopt six fixed pretrained feature extractors: Inception-v3~\citep{szegedy2016rethinking}, ConvNeXt-v2~\citep{Woo2023ConvNeXtVC}, DINOv2~\citep{oquab2024dinov2}, MAE~\citep{he2022masked}, SigLIP2~\citep{tschannen2025siglip}, and CLIP~\citep{radford2021learning}. The FDDA training objective uses the subset $\Psi_{\text{train}} = \{\text{Inception-v3}, \text{MAE}, \text{SigLIP2}\}$, while the remaining three representations are held out for evaluation in all subsequent experiments.

FD values can differ substantially in scale across representation spaces, so directly summing them can cause terms with larger numerical values to dominate the objective. We normalize each representation-specific FD by its stop-gradient value. At iteration $t$, FDDA optimizes
\begin{equation}
    \mathcal{L}_{\text{FDDA}} = \sum_{\psi_k \in \Psi_{\text{train}}} w_k \frac{\operatorname{FD}_{\psi_k}^{(t)}
    }{\operatorname{sg}\left(\operatorname{FD}_{\psi_k}^{(t)}\right)+ c},
    \label{eq:fdda_multi_rep}
\end{equation}
where $\operatorname{sg}(\cdot)$ denotes stop-gradient, $c = 0.01$ is a small constant for numerical stability, and $w_k = 1$ for all representations. The denominator is stop gradient, and serves only as a scale factor during back-propagation, so normalization does not change the gradient direction of each FD term. This reduces the effect of differences in numerical scale across representation spaces.

Estimating the generated feature statistics in Equation~\ref{eq:fdda_multi_rep} requires more samples than are available in a single minibatch. We use an EMA-based estimator for this purpose. For each representation $\psi_k$, the real-image statistics $(\mu_{r,k}, \Sigma_{r,k})$ are precomputed once and kept fixed. The EMA statistics $\mu_{g,k}^{(0)}$ and $M_{g,k}^{(0)}$ are initialized from $50{,}000$ samples drawn from the frozen generator prior to training. At iteration $t$, given a minibatch of 
generation-time latents $\{z_{g,i}^{(t)}\}_{i=1}^{B}$ sampled from the frozen generator, we compute the generated features 
$h_{k,i}^{(t)} = \psi_k(\mathcal{D}_{\phi}(z_{g,i}^{(t)}))$. Let 
$\mu_{b,k}^{(t)} = \frac{1}{B}\sum_{i=1}^{B} h_{k,i}^{(t)}$ and 
$M_{b,k}^{(t)} = \frac{1}{B}\sum_{i=1}^{B} h_{k,i}^{(t)}{h_{k,i}^{(t)}}^{\top}$
denote the batch mean and second moment, respectively. The running generated statistics are updated by
\begin{align*}
    \mu_{g,k}^{(t)}
    &=
    \beta \mu_{g,k}^{(t-1)}
    +
    (1-\beta)\mu_{b,k}^{(t)}, \\
    M_{g,k}^{(t)}
    &=
    \beta M_{g,k}^{(t-1)}
    +
    (1-\beta)M_{b,k}^{(t)}, \\
    \Sigma_{g,k}^{(t)}
    &=
    M_{g,k}^{(t)}
    -
    \mu_{g,k}^{(t)}
    {\mu_{g,k}^{(t)}}^{\top},
    \label{eq:fdda_ema}
\end{align*}
where $\beta=0.999$ is the EMA decay rate. The representation-specific $\operatorname{FD}_{\psi_k}^{(t)}$ in Equation~\ref{eq:fdda_multi_rep} is then computed using the fixed real statistics $(\mu_{r,k}, \Sigma_{r,k})$ and the current generated statistics $(\mu_{g,k}^{(t)}, \Sigma_{g,k}^{(t)})$. EMA statistics carried over from previous iterations are treated as constants during back-propagation, so gradients flow only through features computed from the current minibatch. The latent generator and all representation models remain frozen throughout post-training, and only the decoder parameters $\phi$ are optimized. The EMA estimator avoids storing a large feature population while maintaining running statistics for the FD objective.
\section{Experiments}
\label{sec:experiments}
We conduct extensive experiments to evaluate the \textbf{effectiveness}, \textbf{complementarity}, and \textbf{computational efficiency} of FDDA across diverse autoregressive and flow-based latent generative models. We evaluate FDDA on pretrained models and examine whether it provides further gains after generator-side FD post-training. We next compare its end-to-end post-training cost with generator-side FD post-training. Additional analyses cover longer training, reconstruction performance, and qualitative results. Code and checkpoints will be publicly released to ensure reproducibility.

\subsection{Experimental Setup}
\label{sec:experimental_setup}

\paragraph{Models and Training Setup.}
We evaluate FDDA on a diverse set of pretrained latent generative models. For autoregressive generation, we consider LlamaGen-B and LlamaGen-L~\citep{Sun2024Autoregressive}, GigaTok-S-S~\citep{Xiong2025GigaTokSV}, TiTok-L-32 and TiTok-B-64~\citep{Yu2024AnII}, and VAR-$d16$, VAR-$d20$, and VAR-$d24$~\citep{Tian2024VAR}. For flow-based generation, we consider iMF-B, iMF-L, and iMF-XL~\citep{Geng2025ImprovedMF}.
We apply FDDA to both the original pretrained checkpoints and their generator-side FD-post-trained counterparts from~\citet{Zhang2026Frechetdistributional} for autoregressive models and~\citet{Yang2026RepresentationFL} for flow-based models.
FDDA is implemented in two stages: we first sample generation-time latents from the frozen generator and cache them as a training pool, and then optimize only the decoder on the cached latents using the FDDA objective. Unless otherwise specified, all FDDA models are trained for one epoch. Additional implementation and training details are provided in Appendix~\ref{appendix:experimental details}.

\paragraph{Evaluation Metrics.}
We primarily evaluate generation performance using gFDr$^6$~\citep{Yang2026RepresentationFL}, Fréchet Inception Distance~\citep[gFID;][]{Heusel2017GANsTB}, representation-specific Fr\'echet distances computed in ConvNeXt-v2~\citep{Woo2023ConvNeXtVC}, DINOv2~\citep{oquab2024dinov2}, MAE~\citep{he2022masked}, SigLIP2~\citep{tschannen2025siglip}, and CLIP~\citep{radford2021learning} feature spaces (gFD$_{\text{conv}}$, gFD$_{\text{dino}}$, gFD$_{\text{mae}}$, gFD$_{\text{sig}}$, and gFD$_{\text{clip}}$), and generation Inception Score~\citep[gIS;][]{Salimans2016ImprovedTF}. ConvNeXt-v2, DINOv2, and CLIP are held out from FDDA training and used only for evaluation. We also evaluate reconstruction performance using peak signal-to-noise ratio (PSNR), structural similarity index (SSIM), Fréchet Inception Distance~\citep[rFID;][]{Heusel2017GANsTB}, and perceptual similarity~\citep[LPIPS;][]{Zhang2018TheUE}.

\subsection{Effectiveness and Complementarity of FDDA}
\label{sec:effectiveness}
We evaluate FDDA under two complementary settings. We first apply FDDA to the original pretrained models to isolate decoder-side FD post-training. We then apply FDDA after generator-side FD post-training to test whether further gains remain. Finally, we compare representative FDDA models with a broader set of generative systems.

\begin{table*}[!t]
\centering
\caption{\small{\textbf{FDDA before generator-side FD post-training.} We apply FDDA to the original pretrained latent generative models. All metrics are computed using the same evaluation protocol following~\citet{Yang2026RepresentationFL}. \XSolidBrush and \CheckmarkBold indicate whether each component undergoes FD post-training.}}
\vspace{-2ex}
\label{tab:before generator-side}
\resizebox{\textwidth}{!}{%
\begin{tabular}{l|cc|ccccccc|c}
\toprule
Models & Generator FD & Decoder FD & gFID ({\color{green!60!black}\bfseries$\pmb{\downarrow}$}) & gIS ({\color{green!60!black}\bfseries$\pmb{\uparrow}$}) & gFD$_{\text{conv}}$ ({\color{green!60!black}\bfseries$\pmb{\downarrow}$}) & gFD$_{\text{dino}}$ ({\color{green!60!black}\bfseries$\pmb{\downarrow}$}) & gFD$_{\text{mae}}$ ({\color{green!60!black}\bfseries$\pmb{\downarrow}$}) & gFD$_{\text{sig}}$ ({\color{green!60!black}\bfseries$\pmb{\downarrow}$}) & gFD$_{\text{clip}}$ ({\color{green!60!black}\bfseries$\pmb{\downarrow}$}) &  gFDr$^6$({\color{green!60!black}\bfseries$\pmb{\downarrow}$}) \\
\midrule
\rowcolor{gray!15}
\multicolumn{3}{l}{\textbf{Discrete autoregressive generative models}} \\
\multirow{2}{*}{LlamaGen-B} & \XSolidBrush & \XSolidBrush & 5.17 & 196.1 & 233.91 & 244.12 & 0.528 & 21.03 & 149.44 & 16.48\quad\quad\quad\enspace \\
& \XSolidBrush & \CheckmarkBold & 2.74 & 209.3 & 237.39 & 190.90 & 0.250 & 11.50 & 115.52 & 10.88 \;\;{\color{red} $\downarrow_{34.0\%}$} \\
\arrayrulecolor{gray!25}\cline{1-11}\arrayrulecolor{black}
\multirow{2}{*}{LlamaGen-L} & \XSolidBrush & \XSolidBrush & 4.23 & 290.1 & 113.61 & 121.41 & 0.383 & 12.20 & 97.13 & 10.00\quad\quad\quad\enspace \\
& \XSolidBrush & \CheckmarkBold & 1.87 & 303.5 & 103.27 & 87.98 & 0.109 & 5.19 & 73.07 & 5.60 \;\;{\color{red} $\downarrow_{44.0\%}$} \\
\arrayrulecolor{gray!25}\cline{1-11}\arrayrulecolor{black}
\multirow{2}{*}{GigaTok-S-S} & \XSolidBrush & \XSolidBrush & 4.34 & 268.5 & 135.95 & 182.66 & 0.432 & 15.25 & 112.26 & 12.29\quad\quad\quad\enspace \\
& \XSolidBrush & \CheckmarkBold & 1.68 & 281.6 & 99.96 & 123.43 & 0.169 & 6.32 & 74.75 & 6.59 \;\;{\color{red} $\downarrow_{46.4\%}$} \\
\arrayrulecolor{gray!25}\cline{1-11}\arrayrulecolor{black}
\multirow{2}{*}{TiTok-L-32} & \XSolidBrush & \XSolidBrush & 2.75 & 194.1 & 190.86 & 192.05 & 0.567 & 18.86 & 157.93 & 15.32\quad\quad\quad\enspace \\
& \XSolidBrush & \CheckmarkBold & 1.34 & 202.6 & 181.26 & 116.29 & 0.124 & 5.49 & 70.83 & 6.19 \;\;{\color{red} $\downarrow_{59.6\%}$} \\
\arrayrulecolor{gray!25}\cline{1-11}\arrayrulecolor{black}
\multirow{2}{*}{TiTok-B-64} & \XSolidBrush & \XSolidBrush & 2.44 & 206.6 & 174.01 & 197.82 & 0.509 & 17.36 & 154.73 & 14.57\quad\quad\quad\enspace \\
& \XSolidBrush & \CheckmarkBold & 1.61 & 214.9 & 184.09 & 129.98 & 0.157 & 6.18 & 80.39 & 6.99 \;\;{\color{red} $\downarrow_{52.0\%}$} \\
\arrayrulecolor{gray!25}\cline{1-11}\arrayrulecolor{black}
\multirow{2}{*}{VAR-$d16$} & \XSolidBrush & \XSolidBrush & 3.27 & 280.8 & 148.97 & 150.36 & 0.428 & 13.56 & 104.23 & 11.12\quad\quad\quad\enspace \\
& \XSolidBrush & \CheckmarkBold & 1.62 & 280.1 & 152.45 & 121.49 & 0.175 & 6.68 & 81.14 & 7.03 \;\;{\color{red} $\downarrow_{36.8\%}$} \\
\arrayrulecolor{gray!25}\cline{1-11}\arrayrulecolor{black}
\multirow{2}{*}{VAR-$d20$} & \XSolidBrush & \XSolidBrush & 2.61 & 305.2 & 111.78 & 102.19 & 0.344 & 9.84 & 85.74 & 8.46\quad\quad\quad\enspace \\
& \XSolidBrush & \CheckmarkBold & 1.32 & 296.9 & 124.40 & 82.77 & 0.127 & 4.95 & 65.96 & 5.34 \;\;{\color{red} $\downarrow_{36.9\%}$} \\
\arrayrulecolor{gray!25}\cline{1-11}\arrayrulecolor{black}
\multirow{2}{*}{VAR-$d24$} & \XSolidBrush & \XSolidBrush & 2.11 & 314.7 & 94.78 & 76.34 & 0.286 & 7.37 & 73.93 & 6.79\quad\quad\quad\enspace \\
& \XSolidBrush & \CheckmarkBold & 1.24 & 304.4 & 112.00 & 60.89 & 0.097 & 3.75 & 55.13 & 4.26 \;\;{\color{red} $\downarrow_{37.3\%}$} \\
\midrule
\rowcolor{gray!15}
\multicolumn{3}{l}{\textbf{Flow-match generative models, one-step}} \\
\multirow{2}{*}{iMF-B} & \XSolidBrush & \XSolidBrush & 3.49 & 253.7 & 188.08 & 206.17 & 0.487 & 19.65 & 156.14 & 15.40\quad\quad\quad\enspace \\
& \XSolidBrush & \CheckmarkBold & 1.65 & 268.1 & 115.47 & 144.39 & 0.232 & 8.10 & 105.06 & 8.54 \;\;{\color{red} $\downarrow_{44.5\%}$} \\
\arrayrulecolor{gray!25}\cline{1-11}\arrayrulecolor{black}
\multirow{2}{*}{iMF-L} & \XSolidBrush & \XSolidBrush & 1.88 & 275.4 & 105.53 & 111.56 & 0.282 & 11.30 & 101.18 & 9.11\quad\quad\quad\enspace \\
& \XSolidBrush & \CheckmarkBold & 1.26 & 281.0 & 75.93 & 83.77 & 0.137 & 5.12 & 72.12 & 5.48 \;\;{\color{red} $\downarrow_{39.8\%}$} \\
\arrayrulecolor{gray!25}\cline{1-11}\arrayrulecolor{black}
\multirow{2}{*}{iMF-XL} & \XSolidBrush & \XSolidBrush & 1.81 & 281.9 & 97.57 & 103.01 & 0.261 & 10.37 & 94.58 & 8.44\quad\quad\quad\enspace \\
& \XSolidBrush & \CheckmarkBold & 1.14 & 288.3 & 67.12 & 77.30 & 0.117 & 4.67 & 67.74 & 5.03 \;\;{\color{red} $\downarrow_{40.4\%}$} \\
\bottomrule
\end{tabular}
}
\vspace{-1ex}
\end{table*}

\begin{table*}[!t]
\centering
\caption{\small{\textbf{FDDA after generator-side FD post-training.} We apply FDDA to latent generative models whose generators have already undergone FD post-training. All metrics are computed using the same evaluation protocol following~\citet{Yang2026RepresentationFL}. \XSolidBrush and \CheckmarkBold indicate whether each component undergoes FD post-training.}}
\vspace{-2ex}
\label{tab:after generator-side}
\resizebox{\textwidth}{!}{%
\begin{tabular}{l|cc|ccccccc|c}
\toprule
Models & Generator FD & Decoder FD & gFID ({\color{green!60!black}\bfseries$\pmb{\downarrow}$}) & gIS ({\color{green!60!black}\bfseries$\pmb{\uparrow}$}) & gFD$_{\text{conv}}$ ({\color{green!60!black}\bfseries$\pmb{\downarrow}$}) & gFD$_{\text{dino}}$ ({\color{green!60!black}\bfseries$\pmb{\downarrow}$}) & gFD$_{\text{mae}}$ ({\color{green!60!black}\bfseries$\pmb{\downarrow}$}) & gFD$_{\text{sig}}$ ({\color{green!60!black}\bfseries$\pmb{\downarrow}$}) & gFD$_{\text{clip}}$ ({\color{green!60!black}\bfseries$\pmb{\downarrow}$}) & gFDr$^6$ ({\color{green!60!black}\bfseries$\pmb{\downarrow}$}) \\
\midrule
\rowcolor{gray!15}
\multicolumn{3}{l}{\textbf{Discrete autoregressive generative models}} \\
\multirow{2}{*}{LlamaGen-B} & \CheckmarkBold & \XSolidBrush & 2.45 & 277.9 & 118.37 & 105.41 & 0.174 & 6.07 & 70.55 & 6.33\quad\quad\quad\enspace \\
& \CheckmarkBold & \CheckmarkBold & 2.23 & 275.9 & 134.35 & 95.79 & 0.100 & 4.55 & 61.77 & 5.27 \;\;{\color{red} $\downarrow_{16.7\%}$} \\
\arrayrulecolor{gray!25}\cline{1-11}\arrayrulecolor{black}
\multirow{2}{*}{LlamaGen-L} & \CheckmarkBold & \XSolidBrush & 1.32 & 315.9 & 74.54 & 57.09 & 0.104 & 3.18 & 54.27 & 3.96\quad\quad\quad\enspace \\
& \CheckmarkBold & \CheckmarkBold & 1.34 & 313.5 & 83.61 & 49.14 & 0.052 & 2.40 & 44.71 & 3.18 \;\;{\color{red} $\downarrow_{19.7\%}$} \\
\arrayrulecolor{gray!25}\cline{1-11}\arrayrulecolor{black}
\multirow{2}{*}{GigaTok-S-S} & \CheckmarkBold & \XSolidBrush & 2.52 & 291.6 & 97.96 & 82.97 & 0.219 & 4.45 & 57.20 & 5.34\quad\quad\quad\enspace \\
& \CheckmarkBold & \CheckmarkBold & 1.74 & 296.5 & 82.50 & 72.77 & 0.088 & 2.99 & 46.12 & 3.84 \;\;{\color{red} $\downarrow_{28.1\%}$} \\
\arrayrulecolor{gray!25}\cline{1-11}\arrayrulecolor{black}
\multirow{2}{*}{TiTok-L-32} & \CheckmarkBold & \XSolidBrush & 2.25 & 211.7 & 133.58 & 143.61 & 0.334 & 12.24 & 121.85 & 10.69\quad\quad\quad\enspace \\
& \CheckmarkBold & \CheckmarkBold & 1.35 & 215.1 & 161.52 & 103.08 & 0.105 & 4.65 & 64.74 & 5.49 \;\;{\color{red} $\downarrow_{48.6\%}$} \\
\arrayrulecolor{gray!25}\cline{1-11}\arrayrulecolor{black}
\multirow{2}{*}{TiTok-B-64} & \CheckmarkBold & \XSolidBrush & 1.45 & 238.7 & 102.44 & 132.47 & 0.300 & 9.87 & 111.45 & 9.29\quad\quad\quad\enspace \\
& \CheckmarkBold & \CheckmarkBold & 1.42 & 240.3 & 148.23 & 99.99 & 0.113 & 4.50 & 65.35 & 5.42 \;\;{\color{red} $\downarrow_{41.7\%}$} \\
\arrayrulecolor{gray!25}\cline{1-11}\arrayrulecolor{black}
\multirow{2}{*}{VAR-$d16$} & \CheckmarkBold & \XSolidBrush & 1.85 & 307.9 & 80.75 & 47.20 & 0.154 & 3.30 & 38.19 & 3.65\quad\quad\quad\enspace \\
& \CheckmarkBold & \CheckmarkBold & 1.35 & 307.3 & 77.91 & 45.42 & 0.040 & 1.92 & 34.22 & 2.62 \;\;{\color{red} $\downarrow_{28.2\%}$} \\
\arrayrulecolor{gray!25}\cline{1-11}\arrayrulecolor{black}
\multirow{2}{*}{VAR-$d20$} & \CheckmarkBold & \XSolidBrush & 1.64 & 309.4 & 62.35 & 31.78 & 0.596 & 2.27 & 31.10 & 4.63\quad\quad\quad\enspace \\
& \CheckmarkBold & \CheckmarkBold & 1.08 & 308.1 & 76.47 & 28.58 & 0.026 & 1.28 & 28.03 & 1.98 \;\;{\color{red} $\downarrow_{57.2\%}$} \\
\arrayrulecolor{gray!25}\cline{1-11}\arrayrulecolor{black}
\multirow{2}{*}{VAR-$d24$} & \CheckmarkBold & \XSolidBrush & 1.51 & 320.7 & 55.29 & 26.84 & 0.227 & 1.33 & 25.09 & 2.65\quad\quad\quad\enspace \\
& \CheckmarkBold & \CheckmarkBold & 1.09 & 308.3 & 59.35 & 24.81 & 0.022 & 0.96 & 24.54 & 1.67 \;\;{\color{red} $\downarrow_{37.0\%}$} \\
\midrule
\rowcolor{gray!15}
\multicolumn{3}{l}{\textbf{Flow-matching generative models, one-step}} \\
\multirow{2}{*}{iMF-B} & \CheckmarkBold & \XSolidBrush & 0.86 & 307.1 & 65.97 & 89.05 & 0.154 & 5.46 & 70.35 & 5.57\quad\quad\quad\enspace \\
& \CheckmarkBold & \CheckmarkBold & 0.91 & 304.0 & 74.66 & 80.48 & 0.100 & 3.91 & 59.69 & 4.54 \;\;{\color{red} $\downarrow_{18.5\%}$} \\
\arrayrulecolor{gray!25}\cline{1-11}\arrayrulecolor{black}
\multirow{2}{*}{iMF-L} & \CheckmarkBold & \XSolidBrush & 0.77 & 303.4 & 43.60 & 38.89 & 0.050 & 2.46 & 38.85 & 2.72\quad\quad\quad\enspace \\
& \CheckmarkBold & \CheckmarkBold & 0.80 & 299.1 & 65.13 & 36.92 & 0.035 & 2.05 & 33.72 & 2.43 \;\;{\color{red} $\downarrow_{10.7\%}$} \\
\arrayrulecolor{gray!25}\cline{1-11}\arrayrulecolor{black}
\multirow{2}{*}{iMF-XL} & \CheckmarkBold & \XSolidBrush & 0.76 & 306.7 & 40.34 & 33.85 & 0.043 & 2.21 & 35.74 & 2.46\quad\quad\quad\enspace \\
& \CheckmarkBold & \CheckmarkBold & 0.78 & 303.7 & 57.53 & 32.51 & 0.030 & 1.82 & 31.61 & 2.21 \;\;{\color{red} $\downarrow_{10.2\%}$} \\
\bottomrule
\end{tabular}
}
\vspace{-2ex}
\end{table*}

\paragraph{Before Generator-side FD Post-training.}
Table~\ref{tab:before generator-side} evaluates FDDA on the original pretrained models. Across all 11 autoregressive and flow-based models, FDDA consistently improves both gFID and gFDr$^6$,  with gFDr$^6$ reductions ranging from 34.0-59.6\%. Improvements are also observed across multiple representation spaces, including gFD$_{\text{dino}}$, gFD$_{\text{mae}}$, gFD$_{\text{sig}}$, and gFD$_{\text{clip}}$. For example, on TiTok-L-32, FDDA reduces gFID from 2.75 to 1.34 and gFDr$^6$ from 15.32 to 6.19; on iMF-XL, gFID decreases from 1.81 to 1.14 and gFDr$^6$ from 8.44 to 5.03. These results show that directly adapting the decoder to generation-time latents provides substantial gains without modifying the pretrained latent generator.

\paragraph{After Generator-side FD Post-training.}
We next examine whether FDDA remains effective after the generator has already undergone FD post-training. As shown in Table~\ref{tab:after generator-side}, FDDA further reduces gFDr$^6$ across all 11 models, with improvements ranging from 10.2\% to 57.2\%. Improvements are also observed for gFD$_{\text{dino}}$, gFD$_{\text{mae}}$, gFD$_{\text{sig}}$, and gFD$_{\text{clip}}$, despite the substantially stronger starting points provided by generator-side FD post-training. Notably, gFDr$^6$ decreases from 10.69 to 5.49 on TiTok-L-32 and from 4.63 to 1.98 on VAR-$d20$, corresponding to reductions of 48.6\% and 57.2\%, respectively. Although improvements in individual metrics such as gFID are not uniform in this stronger setting, the consistent gains across the aggregate and multiple representation-space metrics indicate that decoder-side FD post-training remains complementary to generator-side optimization.

\begin{table*}[!t]
\centering
\caption{\small{\textbf{System-level comparison on class-conditional ImageNet $256\times256$.}
NFE denotes the number of function evaluations during inference, and \#Params denotes the number of parameters in the latent generator.
All results use CFG where applicable; $\times2$ indicates the additional function evaluations for CFG.
The best result for each metric is highlighted in bold.
$^{\dagger}$: results reported in Tables~4 and~F.7 of FD-Loss~\citep{Yang2026RepresentationFL}.
$^{\ddagger}$: results measured by us following the evaluation protocol of~\citet{Yang2026RepresentationFL}.
Decoder-side post-training is applied to the generator-side FD-post-trained models (FDDA+).
$^{*}$: CFG is applied only over a time subinterval; NFE is reported as the full-CFG upper bound.
}}
\vspace{-2ex}
\label{tab:system-level comparison}
\resizebox{\textwidth}{!}{%
\begin{tabular}{l|cc|ccccccc|c}
\toprule
Models & NFE & \#params & gFID ({\color{green!60!black}\bfseries$\pmb{\downarrow}$}) & gIS ({\color{green!60!black}\bfseries$\pmb{\uparrow}$}) & gFD$_{\text{conv}}$ ({\color{green!60!black}\bfseries$\pmb{\downarrow}$}) & gFD$_{\text{dino}}$ ({\color{green!60!black}\bfseries$\pmb{\downarrow}$}) & gFD$_{\text{mae}}$ ({\color{green!60!black}\bfseries$\pmb{\downarrow}$}) & gFD$_{\text{sig}}$ ({\color{green!60!black}\bfseries$\pmb{\downarrow}$}) & gFD$_{\text{clip}}$ ({\color{green!60!black}\bfseries$\pmb{\downarrow}$}) & gFDr$^6$ ({\color{green!60!black}\bfseries$\pmb{\downarrow}$}) \\
\midrule
\rowcolor{gray!15}
\multicolumn{3}{l}{\textbf{Without FD post-training}} \\
LlamaGen-L$^{\ddagger}$ & 256$\times$2 & 343M & 4.23 & 290.1 & 113.61 & 121.41 & 0.383 & 12.20 & 97.13 & 10.00 \\
GigaTok-S-S$^{\ddagger}$ & 256$\times$2 & 111M & 4.34 & 268.5 & 135.95 & 182.66 & 0.432 & 15.25 & 112.26 & 12.29 \\
TiTok-L-32$^{\ddagger}$ & 8$\times$2 & 177M & 2.75 & 194.1 & 190.86 & 192.05 & 0.567 & 18.86 & 157.93 & 15.32 \\
VAR-$d24$$^{\ddagger}$ & 10$\times$2 & 1.0B & 2.11 & 314.7 & 94.78 & 76.34 & 0.286 & 7.37 & 73.93 & 6.79 \\
VAR-$d30$$^{\dagger}$ & 10$\times$2 & 2B & 1.97 & 304.6 & 96.57 & 75.35 & 0.29 & 7.19 & 74.61 & 6.70 \\
BAR-L$^{\dagger}$ & 256$\times$2{\tiny$\times$4} & 1.1B & 1.01 & 281.9 & 44.55 & 46.68 & 0.18 & 3.99 & 33.52 & 3.57 \\
SiT-XL/2$^{\dagger}$ & 250$\times$2 & 675M & 2.12 & 256.7 & 114.89 & 111.86 & 0.24 & 9.76 & 99.16 & 8.44 \\
MAR-L$^{\dagger}$ & 256$\times$2{\tiny$\times$100} & 478M & 1.80 & 293.4 & 62.33 & 86.32 & 0.19 & 7.66 & 82.81 & 6.68 \\
FlowAR-H$^{\dagger}$ & 50$\times$2$^{*}$ & 1.9B & 1.68 & 274.1 & 73.68 & 66.42 & 0.20 & 7.71 & 69.99 & 6.13 \\
MAR-H$^{\dagger}$ & 256$\times$2{\tiny$\times$100} & 942M & 1.56 & 299.5 & 54.25 & 70.25 & 0.16 & 6.09 & 72.99 & 5.61 \\
MAR-L, DeTok$^{\dagger}$ & 256$\times$2{\tiny$\times$100} & 478M & 1.39 & 306.2 & 77.56 & 66.14 & 0.19 & 5.79 & 67.94 & 5.49 \\
REG$^{\dagger}$ & 250$\times$2$^{*}$ & 685M & 1.54 & 302.9 & 64.86 & 48.93 & 0.13 & 5.09 & 60.87 & 4.64 \\
SiT-XL/2-REPA$^{\dagger}$ & 250$\times$2$^{*}$ & 675M & 1.42 & 306.1 & 69.36 & 60.62 & 0.17 & 5.97 & 70.90 & 5.45 \\
LightningDiT$^{\dagger}$ & 250$\times$2 & 675M & 1.42 & 294.3 & 62.19 & 53.38 & 0.13 & 5.12 & 57.24 & 4.57 \\
DDT-XL$^{\dagger}$ & 250$\times$2 & 675M & 1.26 & 309.3 & 58.25 & 60.39 & 0.18 & 6.15 & 77.70 & 5.70 \\
REPA-E$^{\dagger}$ & 250$\times$2$^{*}$ & 676M & 1.17 & 298.3 & 72.89 & 34.57 & 0.11 & 3.05 & 35.18 & 3.04 \\
RAE-XL$^{\dagger}$ & 50$\times$2$^{*}$ & 839M & 1.16 & 261.0 & 101.72 & 29.92 & 0.14 & 2.29 & 44.13 & 3.26 \\
Drift-L$^{\dagger}$ & 1 & 463M & 1.53 & 257.2 & 115.37 & 146.88 & 0.28 & 14.59 & 121.01 & 10.92 \\
iMF-XL$^{\ddagger}$ & 1 & 610M & 1.81 & 281.9 & 97.57 & 103.01 & 0.261 & 10.37 & 94.58 & 8.44 \\
iMF-XL$^{\dagger}$ & 2 & 610M & 1.61 & 289.1 & 87.79 & 89.51 & 0.24 & 9.02 & 86.98 & 7.48 \\
\midrule
\rowcolor{gray!15}
\multicolumn{3}{l}{\textbf{Generator-side FD post-training}} \\
LlamaGen-L$^{\ddagger}$ & 256$\times$2 & 343M & 1.32 & 315.9 & 74.54 & 57.09 & 0.104 & 3.18 & 54.27 & 3.96 \\
GigaTok-S-S$^{\ddagger}$ & 256$\times$2 & 111M & 2.52 & 291.6 & 97.96 & 82.97 & 0.219 & 4.45 & 57.20 & 5.34 \\
TiTok-L-32$^{\ddagger}$ & 8$\times$2 & 177M & 2.25 & 211.7 & 133.58 & 143.61 & 0.334 & 12.24 & 121.85 & 10.69 \\
VAR-$d24$$^{\ddagger}$ & 10$\times$2 & 1.0B & 1.51 & \textbf{320.7} & 55.29 & 26.84 & 0.227 & 1.33 & 25.09 & 2.65 \\
iMF-XL$^{\ddagger}$ & 1 & 610M & \textbf{0.76} & 306.7 & \textbf{40.34} & 33.85 & 0.043 & 2.21 & 35.74 & 2.46 \\
\midrule
\rowcolor{gray!15}
\multicolumn{3}{l}{\textbf{Decoder-side FD post-training (Ours)}} \\
LlamaGen-L$^{\ddagger}$ & 256$\times$2 & 343M & 1.34 & 313.5 & 83.61 & 49.14 & 0.052 & 2.40 & 44.71 & 3.18 \\
GigaTok-S-S$^{\ddagger}$ & 256$\times$2 & 111M & 1.74 & 296.5 & 82.50 & 72.77 & 0.088 & 2.99 & 46.12 & 3.84 \\
TiTok-L-32$^{\ddagger}$ & 8$\times$2 & 177M & 1.35 & 215.1 & 161.52 & 103.08 & 0.105 & 4.65 & 64.74 & 5.49 \\
VAR-$d24$$^{\ddagger}$ & 10$\times$2 & 1.0B & 1.09 & 308.3 & 59.35 & \textbf{24.81} & \textbf{0.022} & \textbf{0.96} & \textbf{24.54} & \textbf{1.67} \\
iMF-XL$^{\ddagger}$ & 1 & 610M & 0.78 & 303.7 & 57.53 & 32.51 & 0.030 & 1.82 & 31.61 & 2.21 \\
\bottomrule
\end{tabular}
}
\vspace{-1ex}
\end{table*}

\begin{table*}[t]
\centering
\caption{\small{
\textbf{Computational efficiency of FDDA.} We compare the wall-clock time of generator-side FD post-training and FDDA on a single H100 GPU. All FDDA models are trained for one epoch, with total time including both generation-time latent pool construction and decoder training. Cost reduction is computed as the ratio of generator-side FD training time to the total FDDA time.
}}
\vspace{-2ex}
\label{tab:computational_efficiency}
\resizebox{\textwidth}{!}{%
\begin{tabular}{l|ccc|cccc|c}
\toprule
\multirow{2}{*}{Models}
& \multicolumn{3}{c|}{Generator-side FD}
& \multicolumn{4}{c|}{Decoder-side FD (FDDA)}
& \multirow{2}{*}{Cost Reduction}
\\\hhline{~|---|----|~}
& Epochs & \# Generator Params & Train (h)
& \# Decoder Params & Latent Pool (h) & Train (h) & Total (h) & \\
\midrule
\rowcolor{gray!15}
\multicolumn{9}{l}{\textbf{Discrete autoregressive generative models}} \\
LlamaGen-B
& 1.25 & 111M & 100.6 & 42.5M
& 1.50 & 2.77 & 4.27
& 23.6 \;{\color{red!60!black}$\times$} \\
LlamaGen-L
& 1.25 & 343M & 185.0 & 42.5M
& 4.02 & 2.78 & 6.80
& 27.2 \;{\color{red!60!black}$\times$} \\
GigaTok-S-S
& 1.25 & 111M & 72.2 & 83.8M
& 2.68 & 3.03 & 5.71
& 12.6 \;{\color{red!60!black}$\times$} \\
TiTok-L-32
& 1.25 & 177M & 16.7 & 337.6M
& 0.29 & 3.19 & 3.48
& 4.8 \;{\color{red!60!black}$\times$} \\
TiTok-B-64
& 1.25 & 177M & 16.1 & 118.8M
& 0.60 & 2.75 & 3.35
& 4.8 \;{\color{red!60!black}$\times$} \\
VAR-$d16$
& 1.25 & 310M & 18.9 & 64.7M
& 1.62 & 3.59 & 5.21
& 3.6 \;{\color{red!60!black}$\times$} \\
VAR-$d20$
& 1.25 & 600M & 24.4 & 64.7M
& 2.85 & 3.59 & 6.44
& 3.8 \;{\color{red!60!black}$\times$} \\
VAR-$d24$
& 1.25 & 1.0B & 30.0 & 64.7M
& 4.14 & 3.60 & 7.74
& 3.9 \;{\color{red!60!black}$\times$} \\
\midrule
\rowcolor{gray!15}
\multicolumn{9}{l}{\textbf{Flow-matching generative models, one-step}} \\
iMF-B
& 100 & 89M & 935 & 49.5M
& 0.12 & 3.77 & 3.89
& 240.4 \;{\color{red!60!black}$\times$} \\
iMF-L
& 100 & 409M & 1252 & 49.5M
& 0.51 & 3.76 & 4.27
& 293.2 \;{\color{red!60!black}$\times$} \\
iMF-XL
& 100 & 610M & 1489 & 49.5M
& 0.77 & 3.76 & 4.53
& 328.7 \;{\color{red!60!black}$\times$} \\
\bottomrule
\end{tabular}%
}
\vspace{-1ex}
\end{table*}

\begin{table*}[t]
\centering
\caption{\small{
\textbf{Reconstruction performance before and after FDDA.}
We compare the pretrained and FDDA-adapted decoders using PSNR, SSIM, rFID, and LPIPS. FDDA is trained on generation-time latents, and we report its effect on reconstruction performance here.
}}
\vspace{-2ex}
\label{tab:reconstruction results}
\resizebox{\textwidth}{!}{%
\begin{tabular}{l|cc|cccc|cccc}
\toprule
\multirow{2}{*}{Models} & \multirow{2}{*}{space} & \multirow{2}{*}{\#params}
& \multicolumn{4}{c|}{Pretrained Decoder}
& \multicolumn{4}{c|}{FDDA-adapted Decoder}
\\\hhline{~|~~|----|----}
& & & PSNR({\color{green!60!black}\bfseries$\pmb{\uparrow}$}) & SSIM ({\color{green!60!black}\bfseries$\pmb{\uparrow}$}) & rFID({\color{green!60!black}\bfseries$\pmb{\downarrow}$}) & LPIPS({\color{green!60!black}\bfseries$\pmb{\downarrow}$}) & PSNR({\color{green!60!black}\bfseries$\pmb{\uparrow}$}) & SSIM({\color{green!60!black}\bfseries$\pmb{\uparrow}$}) & rFID({\color{green!60!black}\bfseries$\pmb{\downarrow}$}) & LPIPS({\color{green!60!black}\bfseries$\pmb{\downarrow}$}) \\
\midrule
LlamaGen-B & discrete & 71.9M & 22.53 & 55.3 & 2.20 & 0.138 & 21.73 & 52.9 & 2.82 & 0.154 \\
LlamaGen-L & discrete & 71.9M & 22.53 & 55.3 & 2.20 & 0.138 & 21.93 & 54.0 & 2.55 & 0.150 \\
GigaTok-S-S & discrete & 136.1M & 22.47 & 54.5 & 1.01 & 0.132 & 21.22 & 50.8 & 1.42 & 0.163 \\
TiTok-L-32 & discrete & 641.1M & 17.55 & 35.9 & 2.22 & 0.329 & 16.57 & 33.5 & 1.64 & 0.357 \\
TiTok-B-64 & discrete & 204.8M & 18.74 & 39.8 & 1.69 & 0.258 & 17.84 & 37.4 & 1.66 & 0.284 \\
VAR-$d16$ & discrete & 108.9M & 24.37 & 64.0 & 0.92 & 0.104 & 22.21 & 58.5 & 2.05 & 0.141 \\
VAR-$d20$ & discrete & 108.9M & 24.37 & 64.0 & 0.92 & 0.104 & 19.48 & 57.7 & 2.35 & 0.158 \\
VAR-$d24$ & discrete & 108.9M & 24.37 & 64.0 & 0.92 & 0.104 & 20.43 & 56.9 & 2.38 & 0.148 \\
iMF-B & continuous & 83.7M & 27.61 & 72.5 & 0.75 & 0.068 & 26.65 & 69.7 & 0.92 & 0.081 \\
iMF-L & continuous & 83.7M & 27.61 & 72.5 & 0.75 & 0.068 & 27.06 & 71.0 & 0.86 & 0.076 \\
iMF-XL & continuous & 83.7M & 27.61 & 72.5 & 0.75 & 0.068 & 27.15 & 71.1 & 0.89 & 0.076 \\
\bottomrule
\end{tabular}%
}
\vspace{-2ex}
\end{table*}

\paragraph{System-Level Comparison.}
We compare FDDA with a broader set of class-conditional ImageNet 256${\times}$256 generative models. For systems without FD post-training, we include LlamaGen-L~\citep{Sun2024Autoregressive}, GigaTok-S-S~\citep{Xiong2025GigaTokSV}, TiTok-L-32~\citep{Yu2024AnII}, VAR-$d24$ and VAR-$d30$~\citep{Tian2024VAR}, BAR-L~\citep{Yu2026AutoregressiveIG}, SiT-XL/2~\citep{Ma2024SiTEF}, MAR-L and MAR-H~\citep{Li2024AutoregressiveIG}, FlowAR-H~\citep{Ren2025FlowARSA}, MAR-L with DeTok~\citep{Yang2025LatentDM}, REG~\citep{Wu2025RepresentationEF}, SiT-XL/2-REPA~\citep{Yu2024RepresentationAF}, LightningDiT~\citep{Yao2025ReconstructionVG}, DDT-XL~\citep{Wang2026DDTDD}, REPA-E~\citep{Leng2025REPAEUV}, RAE-XL~\citep{zheng2026diffusion}, Drift-L~\citep{Deng2026GenerativeMV}, and iMF-XL~\citep{Geng2025ImprovedMF}. We also compare matched LlamaGen-L, GigaTok-S-S, TiTok-L-32, VAR-$d24$, and VAR-$d30$ models after generator-side FD post-training and after further decoder-side post-training with FDDA. As shown in Table~\ref{tab:system-level comparison}, FD post-training lowers gFDr$^6$ for all matched models, and applying FDDA afterward yields further reductions. For VAR-$d24$, gFDr$^6$ decreases from 6.79 without FD post-training to 2.65 after generator-side FD post-training and to \textbf{1.67} after FDDA. This is the lowest gFDr$^6$ among the systems compared in Table~\ref{tab:system-level comparison}.

\subsection{Computational Efficiency of FDDA}
\label{sec:efficiency}
We next compare the end-to-end wall-clock time of decoder-side and generator-side FD post-training. As shown in Table~\ref{tab:computational_efficiency}, FDDA reduces post-training time across all evaluated models, including the cost of constructing the generation-time latent pool. For autoregressive models, the reduction ranges from $3.6$-$23.6\times$. The reduction is substantially larger for flow-based models, reaching $240.4$-$328.7\times$ on iMF~\citep{Geng2025ImprovedMF}.

This efficiency comes from three factors. First, FDDA uses only one epoch of decoder adaptation,\footnote{Following~\citet{Yang2026RepresentationFL}, one epoch corresponds to training on $1.28$ million generated samples; $1.25$ epochs therefore correspond to training on $1.60$ million generated samples.} whereas generator-side FD post-training uses $1.25$ epochs for autoregressive models and $100$ epochs for iMF. Second, FDDA caches generation-time latents before training, so gradients pass only through the decoder. Generator-side optimization must additionally backpropagate through the latent generator and its sampling process, resulting in a deeper backward path. Third, FDDA updates fewer parameters because the decoder is smaller than the latent generator for most models, as shown in Table~\ref{tab:computational_efficiency}. These differences in training epochs, back-propagation depth, and parameter count account for the lower wall-clock cost of FDDA.

\subsection{Additional Analyses}
\label{sec:additional}
\begin{wrapfigure}[11]{r}{0.45\textwidth}
\small
\vspace{-5mm}
\begin{center}
\includegraphics[width=0.45\textwidth]{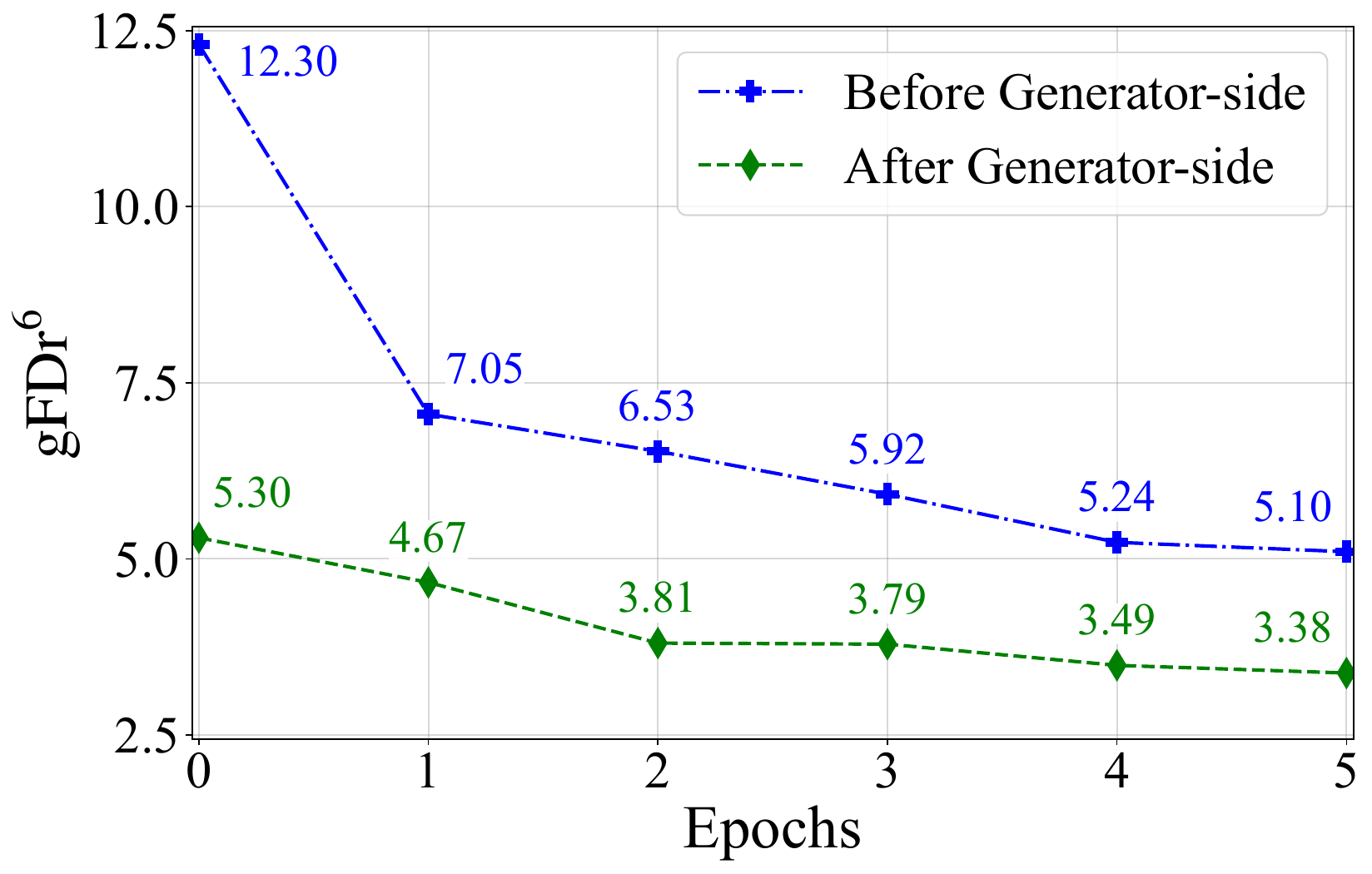}
\vspace{-7mm}
\caption{\small{Effect of FDDA training epochs.}}
\label{fig:training epochs}
\end{center}
\end{wrapfigure}
\paragraph{Effect of Training Epochs} To examine whether longer decoder alignment yields further gains, we train FDDA for five epochs on GigaTok-S-S~\citep{Xiong2025GigaTokSV}, both before and after generator-side FD post-training. As shown in Figure~\ref{fig:training epochs}, gFDr$^6$ continues to decrease throughout all five epochs in both settings. Note that epoch-1 here is not directly comparable to the one-epoch models in Tables~\ref{tab:before generator-side} and~\ref{tab:after generator-side} due to different learning-rate schedules. Extending FDDA from one to five epochs reduces gFDr$^6$ from 6.59 to 5.10 before generator-side post-training and from 3.84 to 3.38 after it.


\paragraph{Reconstruction Performance.}
Table~\ref{tab:reconstruction results} compares reconstruction performance before and after FDDA using PSNR, SSIM, rFID, and LPIPS. Reconstruction quality generally decreases after decoder adaptation: PSNR and SSIM are lower across all models, while LPIPS increases in all cases; rFID also worsens for most models, although it improves slightly for the TiTok models. This follows from the training objective of FDDA: the decoder is adapted on generation-time latents rather than reconstruction latents. This suggests a trade-off between reconstruction fidelity and adaptation to the generation-time latent distribution. Appendix~\ref{app:decoder-latents} ablates the choice of decoder-side FD post-training latents, showing that generation-time latents are key to the observed gains. Qualitative comparisons are provided in Appendix~\ref{appendix:visual comparison}.

\section{Conclusion}
\label{sec:conclusion}

In this work, we extend Fr\'echet distributional post-training from the generator to the decoder in latent generative models. Motivated by the decoder train-generation mismatch, where a decoder trained on reconstruction latents is ultimately applied to generator-produced latents, we introduce \textbf{Fr\'echet Distributional Decoder Alignment (FDDA)}. FDDA keeps the pretrained latent generator fixed and directly optimizes the decoder on generation-time latents using an FD objective. Across diverse flow-based and autoregressive latent generative models, a single epoch of FDDA consistently improves generation quality from the original pretrained checkpoints while requiring substantially less post-training cost than generator-side FD optimization. FDDA further improve gFDr$^6$ after the generator has already undergone FD post-training, showing that generator-side optimization does not exhaust the benefits of distribution-level post-training. More broadly, our results show that distribution-level post-training need not be confined to the latent generator. Generation quality depends both on the generated latent distribution and on how the decoder maps it to images. By aligning the decoder with generation-time latents, FDDA provides a simple and efficient complement to generator-side post-training.

\printbibliography

\clearpage
\begin{appendices}
\crefalias{section}{appsec}
\crefalias{subsection}{appsec}
\crefalias{subsubsection}{appsec}

\setcounter{equation}{0}
\renewcommand{\theequation}{\thesection.\arabic{equation}}

\Huge{Appendix}

\small
\section{Implementation and Training Details}
\label{appendix:experimental details}

\paragraph{Generation-Time Latent Pools.}
For each generator checkpoint, we sample and cache $1.28$M generation-time latents, balanced across the $1{,}000$ ImageNet classes. These cached latents form the training set for decoder-side FD post-training, so one FDDA epoch corresponds to $1.28$M decoder inputs. For autoregressive models, generated tokens are converted to generation-time latents through the frozen tokenizer codebook before being cached as decoder inputs. For flow-based models, we directly cache the final latent states produced by the generator. In both cases, latent pools are constructed using the checkpoint-specific classifier-free guidance (CFG) and sampling configurations used for evaluation.

\paragraph{Decoder-Side FD Post-Training.}
All FDDA models are trained for one epoch using AdamW with $\beta_1=0.9$, $\beta_2=0.95$, and weight decay $10^{-4}$. We use a batch size of 32, corresponding to $40{,}000$ optimization steps per epoch. The learning rate is linearly warmed up to $10^{-5}$ over the first 500 steps and then decayed to $10^{-6}$ using a cosine schedule over the remaining training steps. We apply gradient clipping with maximum norm $1.0$ and use \texttt{bf16} autocast. All experiments are run on a single NVIDIA H100 GPU with 80GB memory. For the longer-training analysis in Figure~\ref{fig:training epochs}, we independently train GigaTok-S-S~\citep{Xiong2025GigaTokSV} for five epochs using the same optimization hyperparameters and learning-rate schedule. The only change is the total training horizon, from $40{,}000$ to $200{,}000$ steps, so the cosine decay is correspondingly stretched over the longer run.

\paragraph{Reference Statistics.}
We use the ImageNet reference statistics released by~\citet{Yang2026RepresentationFL} on Hugging Face for six feature extractors: Inception-v3~\citep{szegedy2016rethinking}, ConvNeXt-v2~\citep{Woo2023ConvNeXtVC}, DINOv2~\citep{oquab2024dinov2}, MAE~\citep{he2022masked}, SigLIP2~\citep{tschannen2025siglip}, and CLIP~\citep{radford2021learning}. The statistics for Inception-v3, MAE, and SigLIP2 are used during FDDA training, while all six are used for evaluation.

\section{Limitations}
\label{appendix:limitations}

Our study has several limitations. First, although we evaluate FDDA across 11 configurations from autoregressive and flow-based latent generative models, all experiments use class-conditional ImageNet generation at $256\times256$ resolution. We have not yet tested FDDA for text-to-image generation, higher-resolution generation, or other data modalities.

Second, FDDA is designed to improve generation by adapting the decoder to generation-time latents. As shown in Table~\ref{tab:reconstruction results}, reconstruction PSNR and SSIM decrease after adaptation across the evaluated models, while LPIPS increases and rFID worsens in most cases. This suggests a trade-off between reconstruction fidelity and adaptation to the latent distribution encountered during generation. For applications that require both reconstruction and generation, an additional reconstruction-preserving objective may be needed during decoder alignment.

Third, FDDA adapts the decoder to a fixed generation-time latent distribution $p_g(z)$ produced by a particular generator checkpoint and inference configuration. Changing the generator or its sampling configuration may change this distribution, in which case the decoder may need to be adapted again.

Finally, FDDA optimizes Fr\'echet distances in a finite set of pretrained representation spaces. Although we additionally evaluate with three representations held out from FDDA training, Fr\'echet objectives capture only first- and second-order feature statistics and may not reflect all aspects of perceptual or semantic generation quality.

\section{Effect of Decoder Post-Training Latents}
\label{app:decoder-latents}

FDDA performs decoder-side FD post-training on generation-time latents $z_g$. To examine the role of this choice, we replace $z_g$ with reconstruction latents $z_r$ obtained by encoding real images, while keeping the decoder-side FD post-training procedure otherwise unchange. We compare the two latent choices on iMF-B, iMF-L, and iMF-XL~\citep{Geng2025ImprovedMF}, starting from either the original pretrained models or their generator-side FD-post-trained counterparts. In all cases, we evaluate generation using samples from the corresponding generator.

Table~\ref{tab:decoder-latents} shows that both latent choices reduce gFDr$^6$ relative to the original pretrained models, but training on $z_g$ produces larger gains. With $z_r$, gFDr$^6$ falls from 15.40 to 12.91 on iMF-B, from 9.11 to 7.13 on iMF-L, and from 8.44 to 6.35 on iMF-XL. The corresponding results with $z_g$ are 8.54, 5.48, and 5.03. Training on $z_g$ also gives lower gFID and lower FD in each of the five other reported feature spaces than training on $z_r$. Notably, gFD$_{\text{conv}}$ increases under $z_r$ adaptation for all three models, whereas it decreases under $z_g$ adaptation.

After generator-side FD post-training, the difference becomes more pronounced in gFDr$^6$. Training on $z_r$ changes it from 5.57 to 5.58 on iMF-B, from 2.72 to 3.19 on iMF-L, and from 2.46 to 2.74 on iMF-XL. Training on $z_g$ instead lowers these values to 4.54, 2.43, and 2.21. It also yields lower FD than $z_r$ adaptation in all five other reported feature spaces for each model. The gains from FDDA therefore depend on the decoder post-training inputs: after generator-side FD post-training, using $z_r$ reverses the gFDr$^6$ benefit, while using the latents encountered at generation time retains it.

\begin{table*}[t]
  \centering
  \caption{\textbf{Effect of decoder post-training latents on iMF.} We compare decoder-side FD post-training using reconstruction latents $z_r$ or generation-time latents $z_g$, starting from either the original pretrained generator or its FD-post-trained counterpart. \XSolidBrush and \CheckmarkBold indicate the absence and presence of FD post-training, respectively, for the component named in each column. All reported metrics evaluate generated images. The percentage beside gFDr$^6$ is the relative change from the row without decoder FD in the same generator setting; downward arrows indicate improvement.}
  \label{tab:decoder-latents}
  \vspace{-2ex}
  \resizebox{\textwidth}{!}{%
  \begin{tabular}{l|ccc|ccccccc|c}
  \toprule
  Models & Generator FD & Decoder FD & Decoder latents
    & gFID ({\color{green!60!black}\bfseries$\pmb{\downarrow}$})
    & gIS ({\color{green!60!black}\bfseries$\pmb{\uparrow}$})
    & gFD$_{\text{conv}}$ ({\color{green!60!black}\bfseries$\pmb{\downarrow}$})
    & gFD$_{\text{dino}}$ ({\color{green!60!black}\bfseries$\pmb{\downarrow}$})
    & gFD$_{\text{mae}}$ ({\color{green!60!black}\bfseries$\pmb{\downarrow}$})
    & gFD$_{\text{sig}}$ ({\color{green!60!black}\bfseries$\pmb{\downarrow}$})
    & gFD$_{\text{clip}}$ ({\color{green!60!black}\bfseries$\pmb{\downarrow}$})
    & gFDr$^6$ ({\color{green!60!black}\bfseries$\pmb{\downarrow}$}) \\
  \midrule
  \multirow{6}{*}{iMF-B} & \XSolidBrush & \XSolidBrush & \Large - & 3.49 & 253.7 & 188.08 & 206.17 & 0.487 & 19.65 & 156.14 & 15.40\quad\quad\quad\enspace \\
   & \XSolidBrush & \CheckmarkBold & \Large $z_r$ & 3.27 & 265.1 & 229.78 & 172.91 & 0.361 & 15.13 & 141.03 & 12.91 \;\;{\color{red} $\downarrow_{16.2\%}$} \\
   & \XSolidBrush & \CheckmarkBold & \Large $z_g$ & 1.65 & 268.1 & 115.47 & 144.39 & 0.232 & 8.10 & 105.06 & 8.54 \;\;{\color{red} $\downarrow_{44.5\%}$} \\ \arrayrulecolor{gray!25}\cline{2-12}\arrayrulecolor{black}
   & \CheckmarkBold & \XSolidBrush & \Large - & \textbf{0.86} & \textbf{307.1} & 65.97 & 89.05 & 0.154 & 5.46 & 70.35 & 5.57\quad\quad\quad\enspace \\
   & \CheckmarkBold & \CheckmarkBold & \Large $z_r$ & 1.08 & 305.5 & 113.59 & 84.02 & 0.158 & 5.29 & 68.01 & 5.58 \;\;{\color{gray} $\uparrow_{0.2\%\phantom{0}}$} \\
   & \CheckmarkBold & \CheckmarkBold & \Large $z_g$ & 0.91 & 304.0 & 74.66 & 80.48 & 0.100 & 3.91 & 59.69 & 4.54 \;\;{\color{red} $\downarrow_{18.5\%}$} \\
  \arrayrulecolor{gray!25}\cline{1-12}\arrayrulecolor{black}
  \multirow{6}{*}{iMF-L} & \XSolidBrush & \XSolidBrush & \Large - & 1.88 & 275.4 & 105.53 & 111.56 & 0.282 & 11.30 & 101.18 & 9.11\quad\quad\quad\enspace \\
   & \XSolidBrush & \CheckmarkBold & \Large $z_r$ & 1.79 & 285.2 & 157.99 & 92.29 & 0.187 & 7.48 & 85.72 & 7.13 \;\;{\color{red} $\downarrow_{21.7\%}$} \\
   & \XSolidBrush & \CheckmarkBold & \Large $z_g$ & 1.26 & 281.0 & 75.93 & 83.77 & 0.137 & 5.12 & 72.12 & 5.48 \;\;{\color{red} $\downarrow_{39.8\%}$} \\\arrayrulecolor{gray!25}\cline{2-12}\arrayrulecolor{black}
   & \CheckmarkBold & \XSolidBrush & \Large - & \textbf{0.77} & \textbf{303.4} & \textbf{43.60} & 38.89 & 0.050 & 2.46 & 38.85 & 2.72\quad\quad\quad\enspace \\
   & \CheckmarkBold & \CheckmarkBold &  \Large $z_r$ & 0.98 & 301.6 & 101.26 & 38.36 & 0.089 & 2.92 & 39.21 & 3.19 \;\;{\color{gray} $\uparrow_{17.3\%}$} \\
   & \CheckmarkBold & \CheckmarkBold & \Large $z_g$ & 0.80 & 299.1 & 65.13 & 36.92 & 0.035 & 2.05 & 33.72 & 2.43 \;\;{\color{red} $\downarrow_{10.7\%}$} \\
  \arrayrulecolor{gray!25}\cline{1-12}\arrayrulecolor{black}
  \multirow{6}{*}{iMF-XL} & \XSolidBrush & \XSolidBrush & \Large - & 1.81 & 281.9 & 97.57 & 103.01 & 0.261 & 10.37 & 94.58 & 8.44\quad\quad\quad\enspace \\
   & \XSolidBrush & \CheckmarkBold & \Large $z_r$ & 1.67 & 289.0 & 141.05 & 85.23 & 0.159 & 6.56 & 76.87 & 6.35 \;\;{\color{red} $\downarrow_{24.8\%}$} \\
   & \XSolidBrush & \CheckmarkBold & \Large $z_g$ & 1.14 & 288.3 & 67.12 & 77.30 & 0.117 & 4.67 & 67.74 & 5.03 \;\;{\color{red} $\downarrow_{40.4\%}$} \\\arrayrulecolor{gray!25}\cline{2-12}\arrayrulecolor{black}
   & \CheckmarkBold & \XSolidBrush & \Large - & \textbf{0.76} & \textbf{306.7} & \textbf{40.34} & 33.85 & 0.043 & 2.21 & 35.74 & 2.46\quad\quad\quad\enspace \\
   & \CheckmarkBold & \CheckmarkBold & \Large $z_r$ & 0.95 & 299.2 & 90.05 & 34.10 & 0.070 & 2.41 & 34.36 & 2.74 \;\;{\color{gray} $\uparrow_{11.4\%}$} \\
   & \CheckmarkBold & \CheckmarkBold & \Large $z_g$ & 0.78 & 303.7 & 57.53 & 32.51 & 0.030 & 1.82 & 31.61 & 2.21 \;\;{\color{red} $\downarrow_{10.2\%}$} \\
  \bottomrule
  \end{tabular}
  }
  \vspace{-2ex}
\end{table*}

\section{Qualitative comparisons}
\label{appendix:visual comparison}
\begin{figure}[!h]
    \centering
    \includegraphics[width=\linewidth]{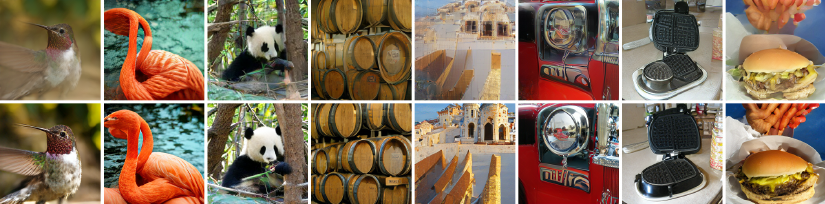}
    \vspace{-4.5ex}
    \caption{\small{\textbf{Generated samples before generator-side FD post-training.} Top: original pretrained model. Bottom: after FDDA.}}
    \label{fig:original_fdda}
    \vspace{-1ex}
\end{figure}

We present qualitative comparisons under three settings. First, we evaluate FDDA before generator-side FD post-training. On TiTok-L-32~\citep{Yu2024AnII}, Figure~\ref{fig:original_fdda} compares the original pretrained model with its FDDA-adapted counterpart, keeping the latent generator fixed. This comparison isolates the effect of decoder-side adaptation.

\begin{figure}[!h]
    \centering
    \includegraphics[width=\linewidth]{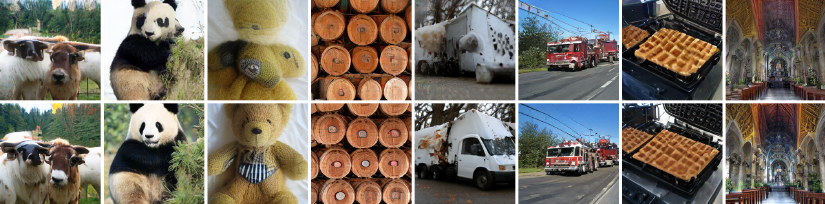}
    \vspace{-4.5ex}
    \caption{\small{\textbf{Generated samples after generator-side FD post-training.} Top: generator-side FD post-training. Bottom: after further applying FDDA.}}
    \label{fig:fdar_fdda_plus}
    \vspace{-1ex}
\end{figure}

Second, we evaluate FDDA after generator-side FD post-training. Figure~\ref{fig:fdar_fdda_plus} compares the generator-side FD post-trained TiTok-L-32 model before and after applying FDDA, showing the additional changes introduced by decoder adaptation when the latent generator has already been post-trained.

\begin{figure}[!h]
    \centering
    \includegraphics[width=\linewidth]{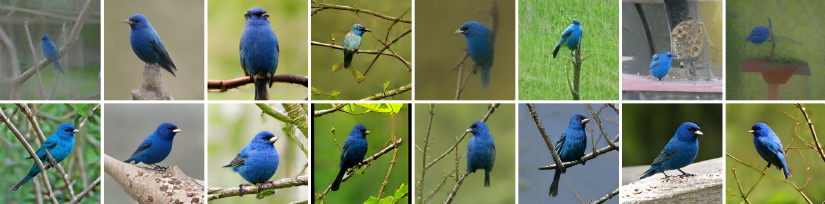}
    \vspace{-4.5ex}
    \caption{\small{\textbf{Class 14: indigo bunting.} Top: original pretrained model. Bottom: generator-side FD post-training + FDDA.}}
    \label{fig:appendix_da_class_014}
    \vspace{-1ex}
\end{figure}

\begin{figure}[!h]
    \centering
    \includegraphics[width=\linewidth]{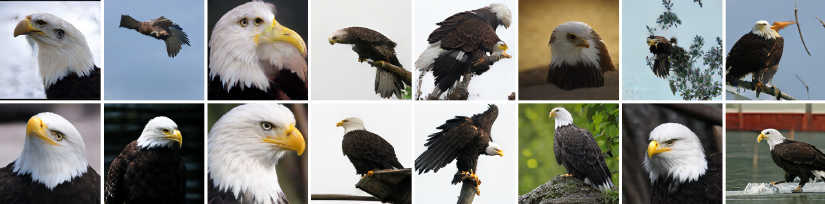}
    \vspace{-4.5ex}
    \caption{\small{\textbf{Class 22: bald eagle.} Top: original pretrained model. Bottom: generator-side FD post-training + FDDA.}}
    \label{fig:appendix_da_class_022}
    \vspace{-1ex}
\end{figure}

\begin{figure}[!h]
    \centering
    \includegraphics[width=\linewidth]{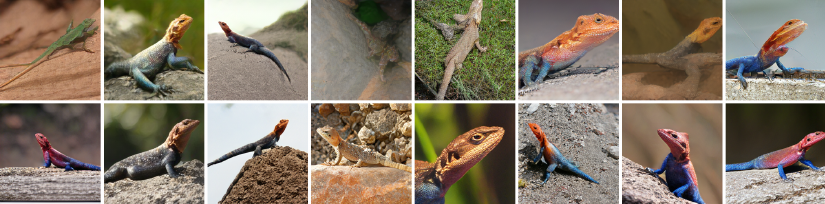}
    \vspace{-4.5ex}
    \caption{\small{\textbf{Class 42: agama.} Top: original pretrained model. Bottom: generator-side FD post-training + FDDA.}}
    \label{fig:appendix_da_class_042}
    \vspace{-1ex}
\end{figure}

\begin{figure}[!h]
    \centering
    \includegraphics[width=\linewidth]{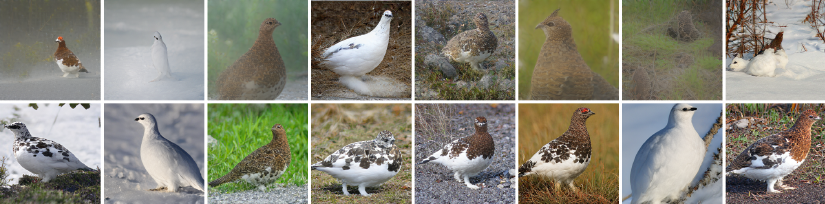}
    \vspace{-4.5ex}
    \caption{\small{\textbf{Class 81: ptarmigan.} Top: original pretrained model. Bottom: generator-side FD post-training + FDDA.}}
    \label{fig:appendix_da_class_081}
    \vspace{-1ex}
\end{figure}

\begin{figure}[!h]
    \centering
    \includegraphics[width=\linewidth]{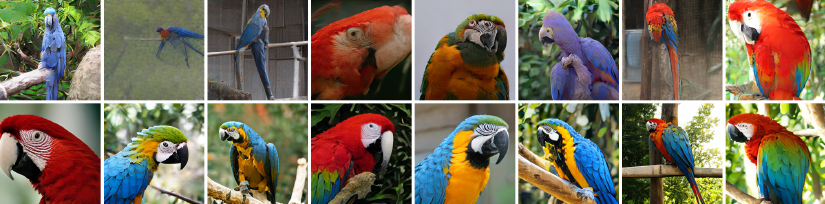}
    \vspace{-4.5ex}
    \caption{\small{\textbf{Class 88: macaw.} Top: original pretrained model. Bottom: generator-side FD post-training + FDDA.}}
    \label{fig:appendix_da_class_088}
    \vspace{-1ex}
\end{figure}

\begin{figure}[!h]
    \centering
    \includegraphics[width=\linewidth]{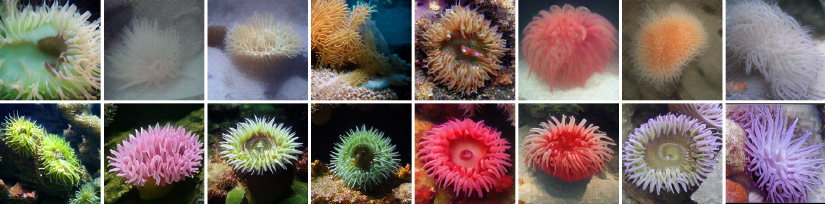}
    \vspace{-4.5ex}
    \caption{\small{\textbf{Class 108: sea anemone.} Top: original pretrained model. Bottom: generator-side FD post-training + FDDA.}}
    \label{fig:appendix_da_class_108}
    \vspace{-1ex}
\end{figure}

\begin{figure}[!h]
    \centering
    \includegraphics[width=\linewidth]{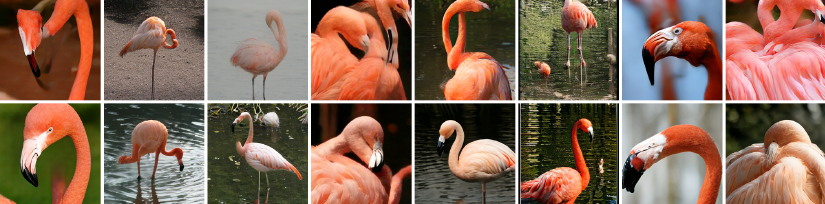}
    \vspace{-4.5ex}
    \caption{\small{\textbf{Class 130: flamingo.} Top: original pretrained model. Bottom: generator-side FD post-training + FDDA.}}
    \label{fig:appendix_da_class_130}
    \vspace{-1ex}
\end{figure}

\begin{figure}[!h]
    \centering
    \includegraphics[width=\linewidth]{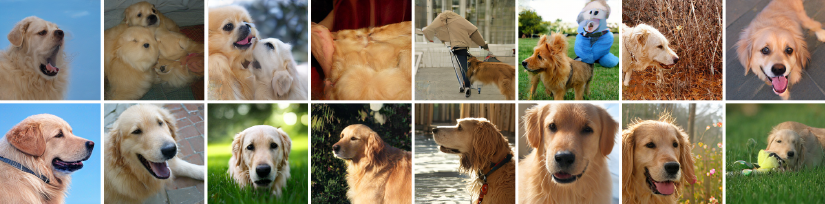}
    \vspace{-4.5ex}
    \caption{\small{\textbf{Class 207: golden retriever.} Top: original pretrained model. Bottom: generator-side FD post-training + FDDA.}}
    \label{fig:appendix_da_class_207}
    \vspace{-1ex}
\end{figure}

\begin{figure}[!h]
    \centering
    \includegraphics[width=\linewidth]{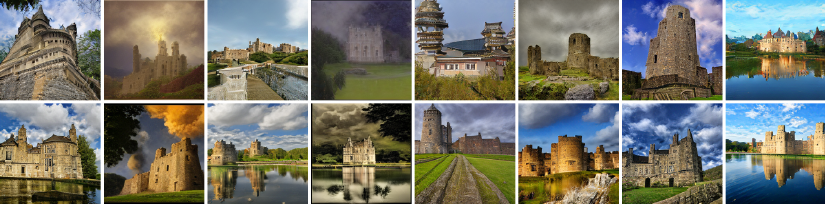}
    \vspace{-4.5ex}
    \caption{\small{\textbf{Class 483: castle.} Top: original pretrained model. Bottom: generator-side FD post-training + FDDA.}}
    \label{fig:appendix_da_class_483}
    \vspace{-1ex}
\end{figure}

\begin{figure}[!h]
    \centering
    \includegraphics[width=\linewidth]{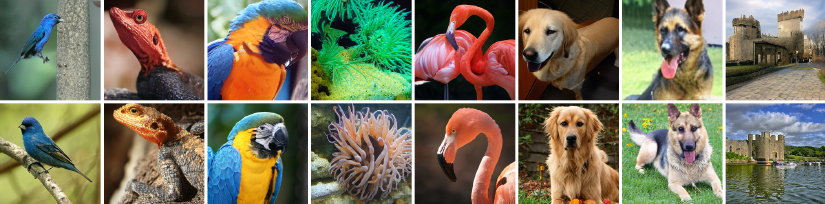}
    \vspace{-4.5ex}
    \caption{\small{\textbf{Generated samples from LlamaGen-L.} Top: original pretrained model. Bottom: generator-side FD post-training + FDDA.}}
    \label{fig:llamagen-l}
    \vspace{-1ex}
\end{figure}

\begin{figure}[!h]
    \centering
    \includegraphics[width=\linewidth]{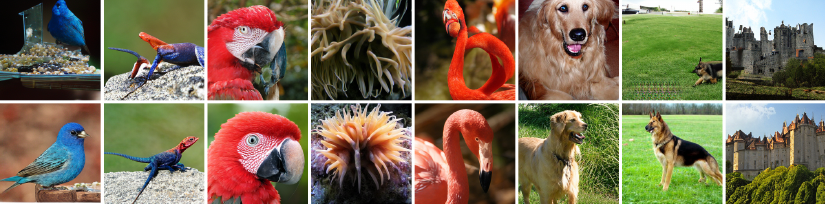}
    \vspace{-4.5ex}
    \caption{\small{\textbf{Generated samples from GigaTok-S-S.} Top: original pretrained model. Bottom: generator-side FD post-training + FDDA.}}
    \label{fig:gigatok}
    \vspace{-1ex}
\end{figure}

\begin{figure}[!h]
    \centering
    \includegraphics[width=\linewidth]{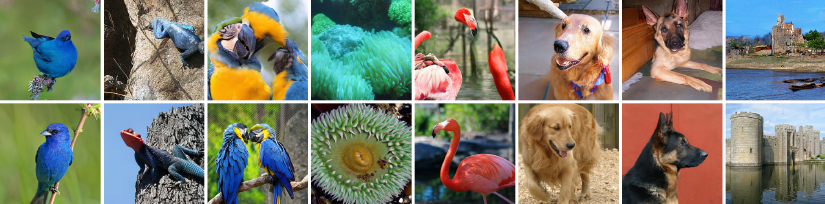}
    \vspace{-4.5ex}
    \caption{\small{\textbf{Generated samples from VAR-$d24$.} Top: original pretrained model. Bottom: generator-side FD post-training + FDDA.}}
    \label{fig:var-d24}
    \vspace{-1ex}
\end{figure}

\begin{figure}[!h]
    \centering
    \includegraphics[width=\linewidth]{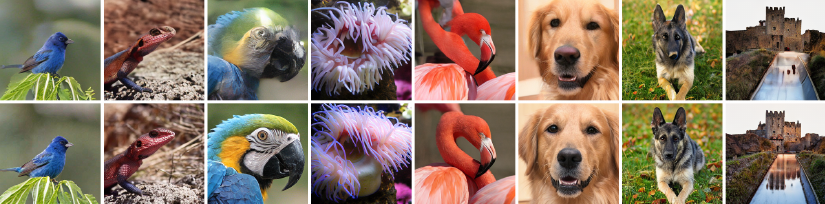}
    \vspace{-4.5ex}
    \caption{\small{\textbf{Generated samples from iMF-XL.} Top: original pretrained model. Bottom: generator-side FD post-training + FDDA.}}
    \label{fig:imf-xl}
    \vspace{-4ex}
\end{figure}

Finally, we compare the original pretrained models with models after both generator-side FD post-training and FDDA. For TiTok-L-32, Figures~\ref{fig:appendix_da_class_014}-\ref{fig:appendix_da_class_483} show examples from 10 ImageNet classes. We also include results on LlamaGen-L~\citep{Sun2024Autoregressive}, GigaTok-S-S~\citep{Xiong2025GigaTokSV}, VAR-$d24$~\citep{Tian2024VAR}, and iMF-XL~\citep{Geng2025ImprovedMF} in Figures~\ref{fig:llamagen-l}-\ref{fig:imf-xl}. Across these examples, the adapted models generally show slightly richer fine-grained details, fewer visible artifacts, and more semantically consistent generations. These visual differences are consistent with the improvements observed in the quantitative evaluation.

\end{appendices}


\end{document}